\documentclass{exam}
\usepackage{amsmath, amssymb}
\usepackage{graphicx}
\usepackage[colorlinks=true, allcolors=blue]{hyperref}

\usepackage[pdftex]{geometry}
\usepackage{enumitem}
\usepackage{bm}
\usepackage{mathtools}
\usepackage[utf8]{inputenc}
\usepackage{graphicx} % For including graphics
\usepackage{subcaption} % For subfigure captions
\usepackage[T1]{fontenc}
\usepackage{amsmath,amsthm,amsfonts,amssymb}
\usepackage{mathtools}
\usepackage{algorithm}
\usepackage{algorithmic}
\usepackage{authblk} % Package for author and affiliation formatting

\usepackage{xcolor}
\theoremstyle{plain}

\theoremstyle{definition}

\theoremstyle{remark}

\newcommand{\R}{\mathbb{R}}

\newcommand{\norm}[1]{\left\lVert#1\right\rVert}
\DeclareMathOperator{\tr}{tr}
\newcommand{\dd}{\ensuremath{\mathrm{d}}}

\begin{document}

\title{Efficient and Scalable Deep Reinforcement Learning for Mean Field Control Games}
% \author{Nianli Peng , Yilin Wang }

% \title{Sample Paper Title}
\author[1]{Nianli Peng$^*$}
\author[1]{Yilin Wang$^*$}
% \author[1,3]{Carol Brown}
\affil[1]{Harvard University}
% \affil[2]{Department of Mathematics, University B}
% \date{December 10, 2024}
\date{}
\maketitle

% \setcitestyle{numbers}

\begin{abstract}
Mean Field Control Games (MFCGs) provide a powerful theoretical framework for analyzing systems of infinitely many interacting agents, blending elements from Mean Field Games (MFGs) and Mean Field Control (MFC). However, solving the coupled Hamilton-Jacobi-Bellman and Fokker-Planck equations that characterize MFCG equilibria remains a significant computational challenge, particularly in high-dimensional or complex environments.

This paper presents a scalable deep Reinforcement Learning (RL) approach to approximate equilibrium solutions of MFCGs. Building on previous works, We reformulate the infinite-agent stochastic control problem as a Markov Decision Process, where each representative agent interacts with the evolving mean field distribution. We use the actor-critic based algorithm from a previous paper \cite{angiuli2024deepreinforcementlearninginfinite} as the baseline and propose several versions of more scalable and efficient algorithms, utilizing techniques including parallel sample collection (batching); mini-batching; target network; proximal policy optimization (PPO); generalized advantage estimation (GAE); and entropy regularization. By leveraging these techniques, we effectively improved the efficiency, scalability, and training stability of the baseline algorithm. 

We evaluate our method on a linear-quadratic benchmark problem, where an analytical solution to the MFCG equilibrium is available. Our results show that some versions of our proposed approach achieve faster convergence and closely approximate the theoretical optimum, outperforming the baseline algorithm by an order of magnitude in sample efficiency. Our work lays the foundation for adapting deep RL to solve more complicated MFCGs closely related to real life, such as large-scale autonomous transportation systems,  multi-firm economic competition, and inter-bank borrowing problems. Our code is available in the Github Repo: 
 \href{https://github.com/InsultedByMaths/6.S890/tree/main}{ https://github.com/InsultedByMaths/6.S890/tree/main}. 
 
 % and information on the project progress (e.g., division of labor, bottleneck, etc.) is documented in appendix \ref{log}

% In addition, we discuss the method’s scalability and potential applicability to more complex multi-agent domains, highlighting the promise of deep RL methods for efficiently navigating the mean-field regime. 

% Our code is available in the Github Repo, and information on the project progress (e.g., division of labor, bottleneck, etc.) is documented in appendix \ref{log}

\footnotetext{* Equal contribution, correspondence to \texttt{nianli\_peng@fas.harvard.edu} and \texttt{yilin\_wang@fas.harvard.edu}}

\end{abstract}

\section{Introduction}\label{sec:introduction}

Multi-agent systems arise in a wide range of engineering, economic, and societal applications, ranging from autonomous vehicle fleets and robotic swarms to large-scale financial markets and energy distribution networks. As these systems grow in complexity and scale, it becomes increasingly challenging to reason about the collective behavior of individual agents and to design effective strategies for coordination, competition, and control. Traditional multi-agent reinforcement learning (MARL) approaches struggle with the exponential growth in complexity as the number of agents increases \cite{busoniu2008comprehensive,zhang2021multi}. When faced with a large or even infinite population of interacting entities, classical game-theoretic, and MARL techniques become computationally prohibitive.

Mean Field Games (MFGs) and Mean Field Control (MFC) offer powerful mathematical frameworks for addressing this scalability challenge. By considering the limit as the number of symmetric agents approaches infinity, MFGs replace the complex interaction of many agents with a representative agent’s interaction against a distribution describing the aggregate state of the population \cite{lasry2007mean,huang2006large}. Under suitable conditions, the solution of an MFG is characterized by a pair of coupled partial differential equations (PDEs): a Hamilton-Jacobi-Bellman (HJB) equation that encodes the optimal control problem for a representative agent, and a Fokker-Planck equation that governs the evolution of the population distribution.

While MFGs capture the notion of individual agents optimizing their own objectives, Mean Field Control (MFC) problems shift the perspective to a central planner seeking to minimize a collective cost for the entire population \cite{carmona2018probabilistic}. Extending these ideas, Mean Field Control Games (MFCGs) bring together elements of both frameworks. In MFCGs, multiple groups or organizers may coordinate internally (as in MFC) while competing externally (as in MFG), resulting in a sophisticated equilibrium concept that must simultaneously satisfy optimal control conditions and distributional fixed points.

However, solving the coupled PDEs that characterize MFGs, MFCs, and MFCGs remains a significant challenge. Analytical solutions are rare and typically limited to stylized problems, while numerical methods that discretize the state space and solve PDEs directly are computationally expensive and scale poorly with dimensionality \cite{achdou2012mean,achdou2020mean}. As a result, there is a pressing need for scalable, data-driven methods capable of approximating equilibrium solutions in complex and high-dimensional settings.

In this paper, we propose a deep Reinforcement Learning (RL) approach to tackle MFCGs at scale. Our method reformulates the infinite-agent stochastic control problem into a Markov Decision Process (MDP), enabling the use of standard RL algorithms. Each representative agent learns a policy that accounts for the evolving mean field distribution, which is itself estimated and updated over the course of training. By adapting an actor-critic framework and introducing batching and target networks, we significantly improve efficiency and stability. This approach circumvents the need to directly solve large PDE systems, leveraging function approximation and sampled trajectories to find near-equilibrium strategies.

We demonstrate our methodology on a linear-quadratic (LQ) benchmark problem that admits an analytical solution \cite{carmonabook}, allowing us to quantitatively compare our results to the theoretical optimum. Our experiments show that the baseline approach (without batching and target networks) struggles with slow convergence and unstable training, whereas our improved algorithm converges an order of magnitude faster and closely approximates the known equilibrium solution. These findings suggest that our RL-based framework is both scalable and effective, opening the door to tackling more complex MFCG problems where no closed-form solutions are available.

The remainder of this paper is structured as follows. In Section~\ref{sec:background}, we review related work on MARL, MFG, MFC, and MFCG, and highlight the gaps that motivate our RL-based approach. Section~\ref{sec:problem} formally presents the MFCG formulation, linking the PDE characterization of equilibrium to a control problem that can be cast as an MDP. Section~\ref{sec:method} describes a reinforcement learning approach towards solving MFCGs. ~\ref{sec:algorithm} detail a baseline algorithm from prior work and our versions of the more efficient algorithm, including the actor-critic framework, batching, mini-batching, the target networks, and commonly used deep RL techniques (PPO, GAE, entropy regularization). We present and analyze experimental details, results, discussions, limitations, and future works in Section~\ref{sec:experiments}. We conclude our work in Section~\ref{sec:discussion}. Some details on formulation and additional ablation experiments are documented in the appendix \ref{appendix} for reference. 

\section{Background and Related Work}\label{sec:background}

In this section, we survey the literature related to multi-agent reinforcement learning, mean field games, mean field control, and their intersection in mean field control games. We highlight how these frameworks address the scalability issues of large-population interactions and identify the gaps that motivate our approach.

% \subsection{Multi-Agent Reinforcement Learning (MARL)}
Classical MARL techniques focus on systems with a finite, and typically small, number of agents \cite{busoniu2008comprehensive,zhang2021multi}. A variety of learning paradigms—such as independent learners, centralized training with decentralized execution, and communication protocols—have been proposed to enable agents to learn coordinated policies. However, as the number of agents grows, these approaches encounter fundamental scalability challenges. The joint state-action space expands combinatorially, and coordination strategies become increasingly complex. Even in simpler settings, finding equilibrium policies or stable solutions can be computationally prohibitive. This exponential blow-up in complexity motivates the need for asymptotic frameworks that simplify analysis and computation.

\subsection{Mean Field Games (MFG)}
Mean Field Games, introduced by Lasry and Lions \cite{lasry2007mean} and further studied by Huang, Caines, and Malham{\'e} \cite{huang2006large}, provide a theoretical foundation for understanding large populations of symmetric agents interacting in a dynamic game. In the limit of an infinite number of agents, the population’s distribution over states (or actions) serves as a proxy for the collective behavior. Each individual (infinitesimal) agent then optimizes its policy against this distribution, rather than explicitly considering every other agent.

Mathematically, MFG equilibria are characterized by a coupled system of PDEs: a Hamilton-Jacobi-Bellman (HJB) equation governing the optimal control for a representative agent, and a Fokker-Planck (FP) equation describing the evolution of the distribution of agents under the aggregate policy. Solving the HJB-FP system yields an equilibrium distribution and a corresponding best-response control. MFG theory has been extensively developed, and under certain conditions, well-posedness and uniqueness results are available \cite{achdou2012mean}.

Yet, from a computational perspective, MFG models still face challenges. Numerical solution of the PDEs can become intractable for high-dimensional state spaces or complex cost structures, limiting their practical applicability \cite{achdou2020mean}.

\subsection{Mean Field Control (MFC)}
While MFGs consider a Nash equilibrium scenario where all agents optimize individually, Mean Field Control (MFC) shifts the viewpoint to that of a central planner or social organizer who aims to minimize an aggregate cost for the entire population \cite{carmona2018probabilistic}. In MFC, the dynamics of the population are still described by McKean-Vlasov-type stochastic differential equations, but the control problem involves choosing a global control policy that influences the distribution’s dynamics to achieve a collective objective.

The solution to MFC problems can also be characterized by PDEs similar to those in MFG, but with different equilibrium conditions. While MFG agents each solve an individual optimal control problem, MFC seeks a centralized solution that aligns the interests of all agents. This perspective simplifies some aspects, as it removes the strategic interaction among agents, but the complexity of solving the underlying PDEs remains significant.

\subsection{Mean Field Control Games (MFCG)}
Mean Field Control Games (MFCGs) combine elements from both MFG and MFC frameworks. Here, multiple groups (or planners) may engage in competitive interactions while coordinating internally \cite{carmona2018probabilistic}. Each group tries to optimize a collective cost subject to the distributional dynamics influenced by other groups. This setting leads to a fixed-point problem that must satisfy both optimal control conditions (as in MFC) and equilibrium constraints (as in MFG).

MFCGs are particularly well-suited to model scenarios where competition and cooperation coexist at different levels of aggregation. For instance, one might consider a landscape of multiple firms each optimizing internal resource allocation (like MFC), while competing for market share against a continuum of other firms (like MFG). Despite their broad applicability, the complexity of MFCGs is even higher than MFGs or MFC alone, as they must capture both the strategic interactions among groups and the optimal control structure within each group.

%\subsection{Gaps in the Literature and Motivation for RL Approaches}
The theoretical developments in MFG, MFC, and MFCG provide a rich understanding of equilibrium conditions for large-scale multi-agent systems. However, their direct solution often relies on PDE-based methods that do not scale well to high-dimensional or otherwise complex environments. Analytical solutions are typically limited to linear-quadratic (LQ) or other highly tractable settings, leaving a large class of practical problems out of reach.

This gap motivates the use of data-driven approximation techniques like reinforcement learning. By framing MFCGs as Markov Decision Processes (MDPs) for representative agents, one can directly apply RL algorithms to approximate optimal policies without explicitly solving the HJB-FP system. Prior works on MFGs have explored RL-based approximations, but scalable and stable algorithms tailored for MFCG settings remain scarce. Our approach aims to address this need by leveraging deep RL methods, combined with batching and target networks, to efficiently approximate MFCG equilibria.

\section{Problem Formulation}\label{sec:problem}

We now formalize the Mean Field Control Game (MFCG) setting. MFCGs combine features of Mean Field Games (MFGs), where agents individually optimize their own strategies against a population distribution, and Mean Field Control (MFC), where a planner optimizes a collective cost for a continuum of agents. In MFCGs, one can think of multiple such planners or groups interacting strategically, leading to equilibrium conditions that must simultaneously account for optimal control and distributional fixed points.

%\subsection{System Dynamics and Agent Population}
Consider an infinite population of agents, each evolving in a continuous-time, continuous-state environment. Let $X_t$ denote the state of a representative agent at time $t$. The evolution of $X_t$ follows a stochastic differential equation (SDE) with McKean-Vlasov dynamics:
\begin{equation}\label{eq:mckean-vlasov}
    dX_t = b(X_t,\mu_t,\alpha(X_t))\, dt + \sigma(X_t,\mu_t,\alpha(X_t))\, dW_t,
\end{equation}
where $\mu_t$ represents the distribution of states across the infinite population at time $t$, $\alpha(\cdot)$ is the control (or policy) applied by the representative agent, $b(\cdot)$ is the drift term, $\sigma(\cdot)$ is the volatility term, and $W_t$ is a standard Brownian motion. The dependence of $b$ and $\sigma$ on $\mu_t$ captures the mean field interaction: each agent’s dynamics depend not just on its own state and action, but also on the collective distribution of the entire population.

%\subsection{Cost Functional and Objective}
Each agent (or planner, in the case of MFC) aims to minimize a discounted infinite-horizon cost functional. For a given policy $\alpha$ and population distribution $\mu$, define the cost:
\begin{equation}\label{eq:cost-functional}
    J(\alpha;\mu) = \mathbb{E}\left[ \int_0^\infty e^{-\beta t} f(X_t,\mu_t,\alpha(X_t)) \, dt \right],
\end{equation}
where $f$ quantifies the running cost, and $\beta > 0$ is a discount factor ensuring integrability. The cost $f$ typically includes terms penalizing control effort (e.g., $\tfrac{1}{2}\|\alpha(X_t)\|^2$) and terms representing ``unsafeness'' or other externalities that depend on the state distribution $\mu_t$. For instance, $f$ might include a component $g(X_t,\mu_t)$ capturing congestion, collision risks, or deviations from desired configurations.

%\subsection{Equilibrium Conditions in MFCGs}
In a Mean Field Game, a Nash equilibrium occurs when no agent can reduce its cost by unilaterally deviating from the equilibrium control, given the equilibrium distribution. Similarly, in Mean Field Control problems, the optimal control is determined by a planner who takes the distribution as given. A Mean Field Control Game extends these concepts by considering that each group or planner might be optimizing a collective cost, potentially in competition or cooperation with others. The equilibrium $(\hat{\alpha},\hat{\mu})$ in an MFCG is a fixed point such that:
\begin{enumerate}
    \item Given the population distribution $\hat{\mu}$, the control $\hat{\alpha}$ solves the stochastic optimal control problem:
    \[
    \hat{\alpha} \in \arg\min_{\alpha} J(\alpha;\hat{\mu}).
    \]
    \item The distribution $\hat{\mu}$ is consistent with the law of the state process under the chosen control $\hat{\alpha}$:
    \[
    \hat{\mu}_t = \mathcal{L}(X_t^{\hat{\alpha},\hat{\mu}}) \quad \text{for all } t \ge 0,
    \]
    where $X_t^{\hat{\alpha},\hat{\mu}}$ evolves according to \eqref{eq:mckean-vlasov} with $\alpha=\hat{\alpha}$ and the distribution fixed at $\hat{\mu}$.
\end{enumerate}

If the problem is time-homogeneous and stationary, we often seek a stationary equilibrium distribution $\hat{\mu}$ and a stationary control $\hat{\alpha}$, leading to a steady-state solution.

%\subsection{Connection to PDE Characterization}
Under suitable regularity conditions, the optimal control problem can be characterized via a Hamilton-Jacobi-Bellman (HJB) equation for the value function associated with the representative agent, and the distribution evolution is governed by a Fokker-Planck (FP) equation. Together, these form a coupled HJB-FP system:
\begin{equation}\label{eq:hjb-fp}
\begin{aligned}
&\text{(HJB): } \partial_t J(x,t) + \min_{\alpha}\biggl\{ \alpha \cdot \nabla_x J(x,t) + f(x,\mu_t,\alpha) \biggr\} + \text{(diffusion terms)}=0,\\[6pt]
&\text{(FP): } \partial_t \mu_t = -\nabla \cdot (\mu_t b(x,\mu_t,\hat{\alpha}(x))) + \frac{1}{2}\nabla \cdot \bigl(\sigma\sigma^T(x,\mu_t,\hat{\alpha}(x)) \nabla_x \mu_t\bigr).
\end{aligned}
\end{equation}
Solving this PDE system directly can be challenging, especially in high dimensions or when $f$, $b$, or $\sigma$ are complex.

%\subsection{Reformulating as a Markov Decision Process (MDP)}
Instead of directly tackling the coupled PDEs, we can use reinforcement learning to approximate the equilibrium solution. By discretizing time and approximating the continuous dynamics with, for instance, an Euler-Maruyama scheme, the representative agent’s optimization problem can be cast as a Markov Decision Process (MDP):
\[
(X_{t},\mu_{t}) \xrightarrow{\alpha_t} (X_{t+1},\mu_{t+1}), \quad \text{with } r_t = -f(X_t,\mu_t,\alpha_t),
\]
and a discount factor $\gamma = e^{-\beta \Delta t}$ for a small time step $\Delta t$. In this MDP viewpoint, $\alpha_t$ represents the agent’s action (control) at state $X_t$, and the transition probabilities reflect the stochastic dynamics influenced by $\mu_t$.

The main challenge is that the distribution $\mu_t$ itself must be learned or approximated. Nevertheless, this RL formulation bypasses the need for explicit PDE solutions. Instead, we can rely on sampling-based methods and function approximation (e.g., neural networks) to iteratively refine both the agent’s policy and the population distribution.

In the next section, we present our deep reinforcement learning methodology, which combines policy optimization with techniques to stabilize and accelerate training, ultimately approximating the MFCG equilibrium solution.

\section{Reinforcement Learning Approach}\label{sec:method}

While the Mean Field Control Game (MFCG) equilibrium can be theoretically characterized by a coupled system of PDEs, directly solving these equations is often infeasible for high-dimensional or complex problems. Instead, we take a data-driven approach by recasting the infinite-agent control problem into a Reinforcement Learning (RL) setting. This viewpoint allows us to leverage function approximation, sampling-based optimization, and established RL techniques to approximate equilibrium solutions.

To apply RL, we discretize time and view the representative agent’s decision-making as a Markov Decision Process (MDP). Suppose we discretize the time horizon into small intervals $\Delta t$. The continuous dynamics given by the McKean-Vlasov equation \eqref{eq:mckean-vlasov} are approximated using an Euler-Maruyama scheme:
\begin{equation}\label{eq:discrete-dynamics}
    X_{t+\Delta t} \approx X_t + b(X_t,\mu_t,\alpha_t)\,\Delta t 
    + \sigma(X_t,\mu_t,\alpha_t)\,\Delta W_t,
\end{equation}
where $\Delta W_t \sim \mathcal{N}(0,\Delta t)$ is a discretized Brownian increment. In this discrete-time setting, the state transitions depend on the chosen control $\alpha_t$ and the distribution $\mu_t$, which influences $b$ and $\sigma$.

The agent observes its state $X_t$, takes an action $A_t = \alpha_t$, and transitions to a new state $X_{t+\Delta t}$. The population distribution $\mu_t$ evolves as well, influencing future transitions. Thus, the representative agent’s problem can be described by an MDP with state space $\mathcal{X}$, action space $\mathcal{A}$, and a transition kernel that depends implicitly on $\mu$.

In the continuous-time formulation, the agent’s objective is to minimize the cost functional:
\[
J(\alpha;\mu) = \mathbb{E}\left[ \int_0^\infty e^{-\beta t} f(X_t,\mu_t,\alpha(X_t)) \, dt \right].
\]
By setting $\gamma = e^{-\beta \Delta t}$, we obtain a discrete-time discounted objective in the RL setting:
\[
J_{\text{RL}}(\alpha;\mu) = \mathbb{E}\left[ \sum_{n=0}^{\infty} \gamma^n f(X_{n\Delta t},\mu_{n\Delta t},\alpha(X_{n\Delta t})) \Delta t \right].
\]
In RL terminology, we define the reward at each time step as:
\[
r_{t} = -f(X_t,\mu_t,A_t)\Delta t,
\]
so that minimizing $J(\alpha;\mu)$ is equivalent to maximizing the expected discounted sum of rewards. This direct mapping allows us to use standard RL algorithms, which are typically formulated in terms of reward maximization.

%\subsection{Approximating the Mean Field Distribution via Score-Matching}\label{subsec: score-matching}
In order to implement the RL approach, we must represent and update the mean field distribution $\mu$ as the system evolves. Directly estimating a probability density over a high-dimensional state space can be challenging. Instead, we adopt a \emph{score-matching} technique \cite{hyvarinen2005}, which has proven effective in generative modeling \cite{song2019generative}.

Score-matching focuses on learning the \emph{score function} $s_\mu(x) \coloneqq \nabla \log p_\mu(x)$, where $p_\mu$ is the density corresponding to $\mu$. Rather than modeling $p_\mu(x)$ directly, we parameterize the score function with a neural network $\Sigma_{\varphi}(x)$ and update $\varphi$ by minimizing an objective that does not require knowing the normalization constant of $p_\mu$. Once we have a good approximation of $s_\mu(x)$, we can generate approximate samples from $\mu$ using Langevin Monte Carlo:
\[
x_{m+1} = x_m + \frac{\epsilon}{2} \Sigma_{\varphi}(x_m) + \sqrt{\epsilon}\,z_m,\quad z_m \sim \mathcal{N}(0,I).
\]
These samples yield an empirical approximation of $\mu$, denoted $\mu_n$ at iteration $n$. As training progresses, $\Sigma_{\varphi}$ and thus $\mu_n$ gradually adjust to reflect the evolving mean field distribution induced by the agent’s policy and environment dynamics.

This score-based approach offers flexibility: it avoids restrictive parametric assumptions about $p_\mu$, can approximate complex distributions, and is compatible with parallel sampling. Though it may introduce additional computational overhead (due to gradient calculations and Langevin steps), this overhead is generally more manageable than attempting a full PDE-based solution.

%\subsection*{Representative Agent Learning and Fixed-Point Condition}
Once the MFCG is modeled as an MDP, the representative agent attempts to find a policy $\Pi_\psi(A|X)$ that, together with a stationary distribution $\hat{\mu}$, forms an equilibrium. In other words, we seek $(\hat{\pi},\hat{\mu})$ such that:
\[
\hat{\pi} \in \arg\max_{\pi} \mathbb{E}\left[ \sum_{n=0}^{\infty} \gamma^n (-f(X_{n},\hat{\mu}_n,A_n))\right],
\]
subject to the consistency condition:
\[
\hat{\mu}_n = \mathcal{L}(X_n^{\hat{\pi},\hat{\mu}}).
\]
Here, $X_n^{\hat{\pi},\hat{\mu}}$ evolves under the chosen policy and the distribution $\hat{\mu}$. Achieving this fixed point involves iteratively updating $\pi_\theta$ and the score network parameters $\varphi$ until stability is reached.

%\subsection*{Leveraging Deep RL Algorithms}
To solve the MDP, we employ deep RL methods that handle continuous action spaces and nonlinear function approximation. We use an actor-critic framework:
\begin{itemize}
    \item \textbf{Actor:} A neural network $\Pi_\psi$ parameterizes the policy, updated via policy gradients.
    \item \textbf{Critic:} A neural network $V_\theta(X)$ estimates the value function, trained using temporal difference methods to reduce variance and guide the policy update.
\end{itemize}

In addition to the actor and critic, we maintain networks for the global and local aspects of the distribution (as represented by the score function $\Sigma_{\varphi}$ and potentially other auxiliary variables capturing local structure in $\mu$). These networks all interact, as the learned distribution influences transitions and rewards, which in turn affect the policy and value function updates.

\subsection{Technical Challenges in Implementation}
Implementing this RL-based framework for MFCGs poses several challenges. Chief among them is the simultaneous learning and mutual dependence of four neural networks: the actor ($\pi_\theta$), the critic ($V_\phi$), and the distributional approximation networks (e.g., the score network $\Sigma_{\varphi}$ and potentially an auxiliary network for local distribution features).

First, the actor and critic form a feedback loop: the actor’s updates depend on value estimates from the critic, while the critic must accurately evaluate returns under the current policy. Next, the distributional approximation must track a moving target since changes in $\pi_\theta$ alter the state visitation frequencies and thus the mean field distribution $\mu$. If we introduce both a global and a local distribution representation, these must also remain consistent with each other and with the evolving environment dynamics.

This intricate web of dependencies can slow convergence and destabilize training. For instance, if the score network $\Sigma_{\varphi}$ lags behind the policy updates, $\mu_n$ may be poorly estimated, causing the actor and critic to learn suboptimal or unstable policies. Conversely, overly aggressive updates to $\Sigma_{\varphi}$ in pursuit of a precise distribution estimate can amplify noise, hindering the policy’s ability to settle into an equilibrium.

Mitigating these issues may require techniques such as:
\begin{itemize}
    \item Careful scheduling of learning rates and update frequencies for different networks.
    \item Using target networks, batching, and parallelization to stabilize learning.
    \item Incorporating regularization or entropy bonuses to prevent overfitting and encourage more robust exploration.
\end{itemize}

These practical challenges underscore the complexity of MFCGs. While our approach lays a foundation for applying deep RL to infinite-agent settings, achieving stable and efficient training still demands careful algorithmic design and parameter tuning, particularly as we scale to richer dynamics and more complex mean field structures.

\section{Proposed Algorithm}\label{sec:algorithm}

% \subsection{Overview}

The proposed algorithm extends the Infinite Horizon Mean Field Control Game Actor-Critic (IH-MFCG-AC) framework \cite{angiuli2024deepreinforcementlearninginfinite}. We propose three editions of the algorithms. 
\begin{itemize}
    \item \textbf{IH-MFCG-AC-B: } Employs parallel sample collection (batching) and added a target network to expedite learning and stabilize the loss curve. B stands for \textbf{B}atching
    \item \textbf{IH-MFCG-AC-M: } Employs parallel sample collection (batching) and adds a target network; At each step, chunks the large batch into mini-batches and performs the learning process within each mini-batch. M stands for \textbf{M}inibatch. 
    \item \textbf{IH-MFCG-AC-DRL: } Based on IH-MFCG-AC-B and employs techniques commonly used in deep reinforcement learning: (1) learn the actor using proximal policy optimization (PPO) \cite{ ppo} instead of REINFORCE \cite{reinforce}; (2) use generalized advantage estimation (GAE) to estimate advantage with variance reduction \cite{gae}; and (3) add entropy regularization to encourage exploration. 
    
\end{itemize}

% First, we adapted the IH-MFCG-AC framework to parallel sample collection (batching) and added a target network to expedite learning and stabilize the loss curve. It is empirically observed that our proposed algorithm converges faster and more stably to the theoretical solution than the baseline IH-MFCG-AC framework. We name this algorithm IH-MFCG-AC-B, where B standards for Batching. Building on IH-MFCG-AC-B, we attempt to employ techniques commonly used in deep reinforcement learning to improve convergence rate and sample efficiency. For learning the actor network, we replaced the REINFORCE algorithm (policy-gradient update rule) with proximal policy optimization (PPO) \cite{reinforce, ppo}. We also make use of Generalized Advantage Estimation (GAE) to approximate the advantage function for variance reduction \cite{gae}. We name this algorithm IH-MFCG-AC-DRL, where DRL stands for deep reinforcement learning. We explore a mini-batch version of IH-MFCG-AC-B with improved trace estimation and reduced Langevin steps, which updates multiple times within a step. We name this algorithm IH-MFCG-AC-M, where M stands for Mini-batching. 

\subsection{Baseline: IH-MFCG-AC}

Algorithm \ref{basealgo} describes the IH-MFCG-AC algorithm.
It uses four neural networks: the actor, the critic, the global score, and the local score. The actor-network learns a policy that selects actions from states and is updated using policy gradient methods, gradually improving the decision-making process. The critic network estimates the value function and is trained by minimizing the temporal difference (TD) error, providing stable feedback for the actor. The global score and local score networks both represent probability distributions as score functions and are trained using a score-matching loss, enabling them to learn the global and local distributions. At each iteration, states are sampled from both the global and local distributions using Langevin Monte Carlo, which updates these distributions as the algorithm proceeds. This combination of actor-critic training and score-based distribution updates leads to convergence toward an equilibrium policy and stationary distributions in the mean field control game. Please refer to the original paper for further details \cite{angiuli2024deepreinforcementlearninginfinite}.

\begin{algorithm}[H]
   \caption{\textbf{IH-MFCG-AC: Infinite Horizon Mean Field Control Game Actor-Critic}}
   \label{basealgo}
\begin{algorithmic}[1] 
    \REQUIRE Initial distribution $\xi$; number of time steps $N \gg 0$; discrete time step size $\Delta t$; neural network learning rates for actor $\rho_\Pi$, critic $\rho_V$, global score $\rho_\Sigma$, and local score $\rho_{\widetilde{\Sigma}}$; Langevin dynamics step size $\epsilon$.
    \STATE Initialize neural networks:\\
    \textbf{Actor} $\Pi_{\psi_0}: \R^d \to \mathcal{P}(\R^k)$\\
    \textbf{Critic} $V_{\theta_0}: \R^d \to \R$\\
    \textbf{Global Score} $\Sigma_{\varphi_0}: \R^d \to \R^d$\\
    {\textbf{Local Score} $\widetilde{\Sigma}_{\xi_0}: \R^d \to \R^d$}\vspace{0.2cm}
    \STATE Sample $X_{t_0} \sim \xi$\vspace{0.2cm}
   \FOR{$n=0,\dots,N-1$}\vspace{0.2cm}
      \STATE Compute score loss for $\Sigma$:
      $\quad L_\Sigma (\varphi_n) = \tr\left( \nabla_x \Sigma_{\varphi_n}(X_{t_n}) \right) + \frac{1}{2}\norm{\Sigma_{\varphi_n}(X_{t_n})}_2^2$\vspace{0.2cm}
      
      \STATE Update $\Sigma$ with Adam:
      $\quad \varphi_{n+1} = \varphi_n -\rho_\Sigma \nabla_{\varphi} L_\Sigma (\varphi_n)$\vspace{0.2cm}

      {\STATE Compute score loss for $\widetilde{\Sigma}$:
      $\quad L_{\widetilde{\Sigma}} (\xi_n) = \tr\left( \nabla_x \widetilde{\Sigma}_{\xi_n}(X_{t_n}) \right) + \frac{1}{2}\norm{\widetilde{\Sigma}_{\xi_n}(X_{t_n})}_2^2$\vspace{0.2cm}

      \STATE Update $\widetilde{\Sigma}$ with Adam:
      $\quad \xi_{n+1} = \xi_n -\rho_{\widetilde{\Sigma}} \nabla_{\xi} L_{\widetilde{\Sigma}} (\xi_n)$}\vspace{0.2cm}
      
      \STATE Generate mean field samples $S_{t_n} = \left(S_{t_n}^{(1)}, S_{t_n}^{(2)}, \dots, S_{t_n}^{(k)}\right)$ from $\Sigma_{\varphi_{n+1}}$ {and $\widetilde{S}_{t_n} = \left(\widetilde{S}_{t_n}^{(1)}, \widetilde{S}_{t_n}^{(2)}, \dots, \widetilde{S}_{t_n}^{(k)}\right)$ from $\widetilde{\Sigma}_{\xi_{n+1}}$} using Langevin dynamics with step size $\epsilon$ and compute $\overline{\mu}_{S_{t_n}} \coloneqq \frac{1}{k} \sum_{i=1}^k \delta_{S_{t_n}^{(i)}}$ {and $\overline{\mu}_{\widetilde{S}_{t_n}} \coloneqq \frac{1}{k} \sum_{i=1}^k \delta_{\widetilde{S}_{t_n}^{(i)}}$}.\vspace{0.2cm}

      \STATE Sample action:
      $\quad A_{t_n} \sim \Pi_{\psi_n}(\cdot \mid X_{t_n})$\vspace{0.2cm}
      
      {\STATE Observe reward from the environment:
      $\quad r_{n+1} = -f(X_{t_n}, \overline{\mu}_{S_{t_n}}, \overline{\mu}_{\widetilde{S}_{t_n}} A_{t_n}) \Delta t$}\vspace{0.2cm}
      
      \STATE Observe the next state from the environment:
        $$\quad X_{t_{n+1}} = b(X_{t_n}, \overline{\mu}_{S_{t_n}}, A_{t_n})\Delta t + \sigma(X_{t_n}, \overline{\mu}_{S_{t_n}}, A_{t_n}) \sqrt{\Delta t} \,z_n, \qquad z_n \sim \mathcal{N}(0,1)$$\vspace{0.1cm}

      \STATE Compute TD target:
      $\quad y_{n+1} = r_{n+1} + e^{-\beta \Delta t} V_{\theta_n}(X_{t_{n+1}})$\vspace{0.1cm}
      
      \STATE Compute TD error:
      $\quad \delta_{\theta_n} =y_{n+1} - V_{\theta_n}(X_{t_n})$\vspace{0.2cm}
      
      \STATE Compute critic loss:
      $\quad L_V(\theta_n) = \delta_{\theta_n}^2$\vspace{0.2cm}
      
      \STATE Update critic with Adam:
      $\quad \theta_{n+1} = \theta_n - \rho_V \nabla_{\theta} L_V(\theta_n)$\vspace{0.2cm}
      
      \STATE Compute actor loss:
      $\quad L_{\Pi}(\psi_n) = -\delta_{\theta_n} \log \Pi_{\psi_n}(A_{t_n} \mid X_{t_n})$\vspace{0.2cm}
      
      \STATE Update actor with Adam:
      $\quad \psi_{n+1} = \psi_n - \rho_{\Pi} \nabla_{\psi} L_\Pi(\psi_n)$\vspace{0.2cm}
   \ENDFOR \vspace{0.2cm}
   \RETURN $(\Pi_{\psi_N}, \Sigma_{\varphi_N}, \widetilde{\Sigma}_{\xi_N})$
\end{algorithmic}
\end{algorithm} 

\subsection{Parallel Sampling and Target Network: IH-MFCG-AC-B}

Algorithm \ref{algo: ihmfcgac} shows the IH-MFCG-AC-B algorithm, with the differences with Algorithm \ref{basealgo} highlighted in {\color{red}{red}}. It introduces two key modifications. First, it employs {parallel sampling} (i.e., batching), meaning that multiple samples of states, actions, and rewards are collected in parallel at each iteration, rather than one at a time. At each update step, we average the loss from each sample. This approach enables us to reduce the variance in the updates at almost no cost of additional training latency due to its effective utilization of GPU acceleration. It is particularly efficient for vectorized environments, and should work well with any environments where it is fast to compute responses given action. Second, we use a {target network} for the critic, which is a copy of the critic’s parameters updated more slowly and thus provides a more stable and less noisy target for the value estimation. Together, these changes increase the stability and scalability of the learning process, making it more robust and less sensitive to randomness (i.e., ``bad luck" ).

\begin{algorithm}[H]
   \caption{\textbf{IH-MFCG-AC-B: IH-MFCG-AC algorithm with Batching and Target Network}}
   \label{algo: ihmfcgac}
\begin{algorithmic}[1] 
    \REQUIRE Initial distribution $\xi$; number of time steps $N \gg 0$; discrete time step size $\Delta t$; neural network learning rates for actor $\rho_\Pi$, critic $\rho_V$, global score $\rho_\Sigma$, and local score $\rho_{\widetilde{\Sigma}}$; Langevin dynamics step size $\epsilon$; {\color{red}batch size $B$}
    \STATE Initialize neural networks:\\
    \textbf{Actor} $\Pi_{\psi_0}: \R^d \to \mathcal{P}(\R^k)$\\
    \textbf{Critic} $V_{\theta_0}: \R^d \to \R$\\
    {\color{red} \textbf{Target} $T_{\theta_0}: \R^d \to \R$, same as Critic} \\
    \textbf{Global Score} $\Sigma_{\varphi_0}: \R^d \to \R^d$\\
    {\textbf{Local Score} $\widetilde{\Sigma}_{\xi_0}: \R^d \to \R^d$}\vspace{0.2cm}
    \STATE {\color{red}Sample $X_{t_0} = \{X_{t_0}^1, X_{t_0}^2, \cdots, X_{t_0}^B \}$, where $X_{t_0}^i \sim \xi$}\vspace{0.2cm}
   \FOR{$n=0,\dots,N-1$}\vspace{0.2cm}
      \STATE Compute score loss for $\Sigma$:
      $\quad L_\Sigma (\varphi_n) = {\color{red} \frac{1}{B}\sum_{i=1}^B}\tr\left( \nabla_x \Sigma_{\varphi_n}(X_{t_n}^{\color{red} i}) \right) + \frac{1}{2}\norm{\Sigma_{\varphi_n}(X_{t_n}^{\color{red} i})}_2^2$\vspace{0.2cm}
      
      \STATE Update $\Sigma$ with Adam:
      $\quad \varphi_{n+1} = \varphi_n -\rho_\Sigma \nabla_{\varphi} L_\Sigma (\varphi_n)$\vspace{0.2cm}

      {\STATE Compute score loss for $\widetilde{\Sigma}$:
      $\quad L_{\widetilde{\Sigma}} (\xi_n) = {\color{red}\frac{1}{B} \sum_{i=1}^B}\tr\left( \nabla_x \widetilde{\Sigma}_{\xi_n}(X_{t_n}^{\color{red} i}) \right) + \frac{1}{2}\norm{\widetilde{\Sigma}_{\xi_n}(X_{t_n}^{\color{red} i})}_2^2$\vspace{0.2cm}

      \STATE Update $\widetilde{\Sigma}$ with Adam:
      $\quad \xi_{n+1} = \xi_n -\rho_{\widetilde{\Sigma}} \nabla_{\xi} L_{\widetilde{\Sigma}} (\xi_n)$}\vspace{0.2cm}
      
      \STATE Generate mean field samples $S_{t_n} = \left(S_{t_n}^{(1)}, S_{t_n}^{(2)}, \dots, S_{t_n}^{(k)}\right)$ from $\Sigma_{\varphi_{n+1}}$ {and $\widetilde{S}_{t_n} = \left(\widetilde{S}_{t_n}^{(1)}, \widetilde{S}_{t_n}^{(2)}, \dots, \widetilde{S}_{t_n}^{(k)}\right)$ from $\widetilde{\Sigma}_{\xi_{n+1}}$} using Langevin dynamics with step size $\epsilon$ and compute $\overline{\mu}_{S_{t_n}} \coloneqq \frac{1}{k} \sum_{i=1}^k \delta_{S_{t_n}^{(i)}}$ {and $\overline{\mu}_{\widetilde{S}_{t_n}} \coloneqq \frac{1}{k} \sum_{i=1}^k \delta_{\widetilde{S}_{t_n}^{(i)}}$}.\vspace{0.2cm}

      \STATE Sample action {\color{red} for i = 1, 2, ..., $B$ (vectorized)}:
      $\quad A_{t_n}^{\color{red} i})  \sim \Pi_{\psi_n}(\cdot \mid X_{t_n}^{\color{red} i}))$\vspace{0.2cm}
      
      {\STATE Observe reward from environment {\color{red} for i = 1, 2, ..., $B$}:
      $\quad r_{n+1}^{\color{red} i} = -f(X_{t_n}^{\color{red} i}, \overline{\mu}_{S_{t_n}}, \overline{\mu}_{\widetilde{S}_{t_n}}, A_{t_n}^{\color{red} i}) \Delta t$}\vspace{0.2cm}
      
      \STATE Observe next state from environment {\color{red} for i = 1, 2, ..., $B$ (vectorized)}:\\
        $\quad X_{t_{n+1}}^{\color{red} i} = b(X_{t_n}^{\color{red} i}, \overline{\mu}_{S_{t_n}}, A_{t_n}^{\color{red} i})\Delta t + \sigma(X_{t_n}^{\color{red} i}, \overline{\mu}_{S_{t_n}}, A_{t_n}^{\color{red} i}) \sqrt{\Delta t} \,z_n, \qquad z_n \sim \mathcal{N}(0,1)$\vspace{0.2cm}

      \STATE Compute TD target {\color{red} for i = 1, 2, ..., $B$ (vectorized)}:
      $\quad y_{n+1}^{\color{red} i} = r_{n+1}^{\color{red} i} + e^{-\beta \Delta t} {\color{red}{T_{\theta_n}(X_{t_{n+1}}^i)}}$\vspace{0.2cm}
      
      \STATE Compute TD error  {\color{red} for i = 1, 2, ..., $B$ (vectorized)}:
      $\quad \delta_{\theta_n}^{\color{red} i} =y_{n+1}^{\color{red} i} - V_{\theta_n}(X_{t_n}^{\color{red} i})$\vspace{0.2cm}
      
      \STATE Compute critic loss:
      $\quad L_V(\theta_n) = {\color{red} \frac{1}{B}\sum_{i=1}^B}(\delta_{\theta_n}^{\color{red} i})^2$\vspace{0.2cm}
      
      \STATE Update critic with Adam:
      $\quad \theta_{n+1} = \theta_n - \rho_V \nabla_{\theta} L_V(\theta_n)$\vspace{0.2cm}
      
      \STATE Compute actor loss:
      $\quad L_{\Pi}(\psi_n) = {\color{red} \frac{1}{B}\sum_{i=1}^B}-\delta_{\theta_n}^{\color{red} i} \log \Pi_{\psi_n}(A_{t_n}^{\color{red} i} \mid X_{t_n})$\vspace{0.2cm}
      
      \STATE Update actor with Adam:
      $\quad \psi_{n+1} = \psi_n - \rho_{\Pi} \nabla_{\psi} L_\Pi(\psi_n)$\vspace{0.2cm}

      \STATE {\color{red} 
      $T_{n+1} \leftarrow V_{n+1}$ if n divisible by 200, else $T_{n+1} \leftarrow T_n$}
   \ENDFOR \vspace{0.2cm}
   \RETURN $(\Pi_{\psi_N}, \Sigma_{\varphi_N}, \widetilde{\Sigma}_{\xi_N})$
\end{algorithmic}
\end{algorithm} 

\subsection{Mini-Batching: IH-MFCG-AC-M}

Algorithm~\ref{algo: minibatch} shows the IH-MFCG-AC-M algorithm. The difference from the IH-MFCG-AC-B algorithm is highlighted in {\color{violet} purple}. Here, at every step, instead of doing the entire learning process once for all samples in a vectorized manner, we first shuffle the states, partition them into mini-batches, and then perform the update procedures for each mini-batch. This approach can improve training stability and better utilize computational resources. For instance, if one has access to large GPU memory, increasing the batch size can lead to higher variance reduction in stochastic gradients. However, simply processing one massive batch at each step might not fully exploit the potential computational parallelism or may lead to slower per-iteration updates. By chunking the batch into mini-batches, the algorithm can perform multiple gradient updates per iteration at a finer granularity, effectively combining the variance reduction benefits of large batches with more frequent parameter updates.

Introducing mini-batching also allows for a more balanced computational load at each iteration. Instead of waiting for the entire large batch to be processed, updates can be made incrementally, potentially stabilizing training dynamics and facilitating faster convergence. In practice, this modification translates into improved performance and reduced training times, especially when combined with optimized parallel computing strategies.

In addition to mini-batching, we make several other improvements over the baseline IH-MFCG-AC-B algorithm:

\paragraph{Improved Trace Estimation for Score Matching.}  
In the previous version, computing the divergence term required by the score matching objective was done by explicitly looping over each dimension and using automatic differentiation to compute partial derivatives. This approach is computationally expensive and does not scale well. To address this, we employ Hutchinson’s trace estimator, which replaces the explicit computation of the divergence with a stochastic approximation. If $\Sigma_\varphi(x)$ is the score approximation at state $x$, and $z \sim \mathcal{N}(0,I)$, then:
\[
\mathrm{div}(\Sigma_\varphi)(x) = \mathbb{E}_z \bigl[z^\top \nabla_x \Sigma_\varphi(x)\bigr] \approx z^\top \nabla_x \Sigma_\varphi(x).
\]
In practice, a single $z$ suffices. This avoids explicit loops over the state dimension, leading to a significant computational speed-up. The loss function can thus be computed more efficiently, enabling larger batch sizes and more complex function approximators without incurring a large computational penalty.

\paragraph{Reduced Frequency of Langevin Updates.}  
Previously, Langevin dynamics were performed at each training iteration to generate samples from the learned score distributions, ensuring proper exploration of the state space and stable estimates of the equilibrium distribution. However, running Langevin updates at every iteration is computationally expensive. In the improved version, we keep the same number of Langevin steps per update but reduce how frequently we perform these updates. By doing so, we rely on the improved score estimates—achieved through better trace estimation and more frequent parameter updates via mini-batching—to maintain adequate distributional coverage. This reduction in update frequency decreases computational overhead and can lead to faster overall training convergence, without significantly compromising the quality of the learned distributions.

\paragraph{Combined Effect of Improvements.}  
Taken together, the introduction of mini-batching, the use of Hutchinson’s trace estimator for the divergence, and the reduction in Langevin steps each contribute to making the IH-MFCG-AC-M algorithm more computationally efficient. Mini-batching allows multiple updates within each iteration, increasing training throughput. The Hutchinson’s estimator streamlines the gradient computations for score matching, making it feasible to use larger models or larger batch sizes. Reducing Langevin steps further decreases computational costs and can lead to faster training iterations.

These improvements, while conceptually simple, jointly enhance the scalability and efficiency of the IH-MFCG-AC algorithm, making it more practical for large-scale problems or more complex function approximators. In the following sections, we will detail the experimental setup and present empirical evidence demonstrating these performance gains.

\begin{algorithm}[H]
   \caption{\textbf{IH-MFCG-AC-M: IH-MFCG-AC Algorithm with Mini-Batching}}
   \label{algo: minibatch}
\begin{algorithmic}[1] 
    \REQUIRE Initial distribution $\xi$; number of time steps $N \gg 0$; discrete time step size $\Delta t$; neural network learning rates for actor $\rho_\Pi$, critic $\rho_V$, global score $\rho_\Sigma$, and local score $\rho_{\widetilde{\Sigma}}$; Langevin dynamics step size $\epsilon$; {\color{red}batch size $B$, } {\color{violet}minibatch size $b$ ($B$ divisible by $b$), number mini-batches $C= B / b$}
    \STATE Initialize neural networks:\\
    \textbf{Actor} $\Pi_{\psi_0}: \R^d \to \mathcal{P}(\R^k)$\\
    \textbf{Critic} $V_{\theta_0}: \R^d \to \R$\\
    {\color{red} \textbf{Target} $T_{\theta_0}: \R^d \to \R$, same as Critic} \\
    \textbf{Global Score} $\Sigma_{\varphi_0}: \R^d \to \R^d$\\
    {\textbf{Local Score} $\widetilde{\Sigma}_{\xi_0}: \R^d \to \R^d$}\vspace{0.2cm}
    \STATE {\color{red}Sample $X_{t_0} = \{X_{t_0}^1, X_{t_0}^2, \cdots, X_{t_0}^B \}$, where $X_{t_0}^i \sim \xi$}\vspace{0.2cm}
   \FOR{$n=0,\dots,N-1$}\vspace{0.2cm}
   \STATE {\color{violet} Randomly Permute $X_{t_n}$ } \vspace{0.2cm}
   \FOR{$m=0,\dots,C-1$}\vspace{0.2cm}

      \STATE Compute score loss for $\Sigma$ using Hutchinson’s trace estimator:
      $\quad L_\Sigma (\varphi_n) = {\color{violet}\frac{1}{b}\sum_{i=m\cdot b +1}^{(m+1)\cdot b } \left( z_i^\top \nabla_x \Sigma_{\varphi_n}(X_{t_n}^{ i}) + \frac{1}{2}\|\Sigma_{\varphi_n}(X_{t_n}^{ i})\|_2^2 \right)}$\vspace{0.2cm}
      
      \STATE Update $\Sigma$ with Adam:
      $\quad \varphi_{n+1} = \varphi_n -\rho_\Sigma \nabla_{\varphi} L_\Sigma (\varphi_n)$\vspace{0.2cm}

      {\STATE Compute score loss for $\widetilde{\Sigma}$ using Hutchinson’s trace estimator:
      $\quad L_{\widetilde{\Sigma}} (\xi_n) = {\color{violet}\frac{1}{b}\sum_{i=m\cdot b +1}^{(m+1)\cdot b } \left( \tilde{z}_i^\top \nabla_x \widetilde{\Sigma}_{\xi_n}(X_{t_n}^{\color{violet} i}) + \frac{1}{2}\|\widetilde{\Sigma}_{\xi_n}(X_{t_n}^{\color{violet} i})\|_2^2 \right)}$\vspace{0.2cm}

      \STATE Update $\widetilde{\Sigma}$ with Adam:
      $\quad \xi_{n+1} = \xi_n -\rho_{\widetilde{\Sigma}} \nabla_{\xi} L_{\widetilde{\Sigma}} (\xi_n)$}\vspace{0.2cm}
      
      \STATE {\color{violet}If n is divisible by 50}, generate mean field samples $S_{t_n} = \left(S_{t_n}^{(1)}, S_{t_n}^{(2)}, \dots, S_{t_n}^{(k)}\right)$ from $\Sigma_{\varphi_{n+1}}$ {and $\widetilde{S}_{t_n} = \left(\widetilde{S}_{t_n}^{(1)}, \widetilde{S}_{t_n}^{(2)}, \dots, \widetilde{S}_{t_n}^{(k)}\right)$ from $\widetilde{\Sigma}_{\xi_{n+1}}$} using Langevin dynamics with step size $\epsilon$ and compute $\overline{\mu}_{S_{t_n}} \coloneqq \frac{1}{k} \sum_{i=1}^k \delta_{S_{t_n}^{(i)}}$ {and $\overline{\mu}_{\widetilde{S}_{t_n}} \coloneqq \frac{1}{k} \sum_{i=1}^k \delta_{\widetilde{S}_{t_n}^{(i)}}$}.\vspace{0.2cm}

      \STATE Sample action {\color{violet} for i = $m\cdot b+1, m\cdot b+2, \cdots, (m+1)\cdot b$ (vectorized)}:
      $\quad A_{t_n}^{\color{red} i})  \sim \Pi_{\psi_n}(\cdot \mid X_{t_n}^{\color{red} i}))$\vspace{0.2cm}
      
      {\STATE Observe reward from environment {\color{violet} for i = $m\cdot b+1, m\cdot b+2, \cdots, (m+1)\cdot b$}:
      $\quad r_{n+1}^{\color{red} i} = -f(X_{t_n}^{\color{red} i}, \overline{\mu}_{S_{t_n}}, \overline{\mu}_{\widetilde{S}_{t_n}}, A_{t_n}^{\color{red} i}) \Delta t$}\vspace{0.2cm}
      
      \STATE Observe next state from environment {\color{violet} for i = $m\cdot b+1, m\cdot b+2, \cdots, (m+1)\cdot b$ (vectorized)}:\\
        $\quad X_{t_{n+1}}^{\color{red} i} = b(X_{t_n}^{\color{red} i}, \overline{\mu}_{S_{t_n}}, A_{t_n}^{\color{red} i})\Delta t + \sigma(X_{t_n}^{\color{red} i}, \overline{\mu}_{S_{t_n}}, A_{t_n}^{\color{red} i}) \sqrt{\Delta t} \,z_n, \qquad z_n \sim \mathcal{N}(0,1)$\vspace{0.2cm}

      \STATE Compute TD target {\color{violet} for i = $m\cdot b+1, m\cdot b+2, \cdots, (m+1)\cdot b$ (vectorized)}:
      $\quad y_{n+1}^{\color{red} i} = r_{n+1}^{\color{red} i} + e^{-\beta \Delta t} {\color{red}{T_{\theta_n}(X_{t_{n+1}}^i)}}$\vspace{0.2cm}
      
      \STATE Compute TD error  {\color{violet} for i = $m\cdot b+1, m\cdot b+2, \cdots, (m+1)\cdot b$ (vectorized)}:
      $\quad \delta_{\theta_n}^{\color{red} i} =y_{n+1}^{\color{red} i} - V_{\theta_n}(X_{t_n}^{\color{red} i})$\vspace{0.2cm}
      
      \STATE Compute critic loss:
      $\quad L_V(\theta_n) = {\color{violet} \frac{1}{b}\sum_{i=m\cdot b +1}^{(m+1)\cdot b }}(\delta_{\theta_n}^{\color{violet} i})^2$\vspace{0.2cm}
      
      \STATE Update critic with Adam:
      $\quad \theta_{n+1} = \theta_n - \rho_V \nabla_{\theta} L_V(\theta_n)$\vspace{0.2cm}
      
      \STATE Compute actor loss:
      $\quad L_{\Pi}(\psi_n) = {\color{violet} \frac{1}{b}\sum_{i=m\cdot b +1}^{(m+1)\cdot b }}-\delta_{\theta_n}^{\color{violet} i} \log \Pi_{\psi_n}(A_{t_n}^{\color{violet} i} \mid X_{t_n})$\vspace{0.2cm}
      
      \STATE Update actor with Adam:
      $\quad \psi_{n+1} = \psi_n - \rho_{\Pi} \nabla_{\psi} L_\Pi(\psi_n)$\vspace{0.2cm}
      \ENDFOR \vspace{0.2cm}

      \STATE {\color{red} 
      $T_{n+1} \leftarrow V_{n+1}$ if n divisible by 200, else $T_{n+1} \leftarrow T_n$}
   \ENDFOR \vspace{0.2cm}
   \RETURN $(\Pi_{\psi_N}, \Sigma_{\varphi_N}, \widetilde{\Sigma}_{\xi_N})$
\end{algorithmic}
\end{algorithm}

\subsection{PPO, GAE, Entropy Regularization: IH-MFCG-AC-DRL}

Algorithm \ref{algo: drl} builds on top of Algorithm \ref{algo: ihmfcgac} by adding components commonly used in deep reinforcement learning. The difference is highlighted in {\color{blue}blue}, and the mathematical formulation is detailed in the algorithm. To provide a brief summary, first, we estimate the advantage function using Generalized Advantage Estimation (GAE) \cite{gae}. This approach reduces variance while maintaining low bias in policy gradient estimation. Note that we introduce a trajectory buffer with a fixed rollout length to store the past trajectories for GAE's calculation, and we only update the actor-critic network per rollout length.  

Secondly, we replaced the actor's update rule with Proximal Policy Optimization (PPO) \cite{ppo}, which restricts the policy update by clipping the probability ratio between old/new actors and improves stability. Note that we have also considered TRPO \cite{trpo} but have decided on PPO for efficiency concerns. Lastly, we introduced entropy regularization in the actor's loss, which encourages exploration by preventing the policy from converging too quickly into a possibly local optimum.

\textbf{Please Continue Next Page}

% \paragraph{Math-Oriented Summary of Changes:}
% Let $r_{t_m}^i(\psi)$ be the probability ratio between the new and old policies, $r_{t_m}^i(\psi)=\frac{\Pi_{\psi}(A_{t_m}^i|X_{t_m}^i)}{\Pi_{\psi_{\text{old}}}(A_{t_m}^i|X_{t_m}^i)}$. The PPO objective modifies the actor loss to:
% \[
% L_{\Pi}(\psi) = \mathbb{E}\left[\min\bigl(r_{t_m}^i(\psi)\hat{A}_{t_m}^i, \text{clip}(r_{t_m}^i(\psi),1-\epsilon_{\text{clip}},1+\epsilon_{\text{clip}})\hat{A}_{t_m}^i\bigr) + c_{\text{ent}}\mathcal{H}(\Pi_{\psi})\right],
% \]
% where $\epsilon_{\text{clip}}$ is the PPO clipping parameter and $c_{\text{ent}}$ is the entropy coefficient. The entropy term $\mathcal{H}(\Pi_{\psi})$ encourages exploration. GAE replaces the advantage estimate $\hat{A}_{t_m}^i$ with:
% \[
% \hat{A}_{t_m}^i = \delta_m^i + \gamma \lambda \hat{A}_{t_{m+1}}^i, \quad \text{where } \delta_m^i = r_{m+1}^i + \gamma V_\theta(X_{t_{m+1}}^i)-V_\theta(X_{t_m}^i),
% \]
% reducing variance while maintaining low bias. These mathematical changes aim for more robust and sample-efficient updates of the policy and value function.

\begin{algorithm}[H]
   \caption{\textbf{IH-MFCG-AC-DRL: PPO with entropy regularization and GAE with Rollouts}}
   \label{algo: drl}
\begin{algorithmic}[1] 
    \REQUIRE Initial distribution $\xi$; number of time steps $N \gg 0$; discrete time step size $\Delta t$; neural network learning rates for actor $\rho_\Pi$, critic $\rho_V$, global score $\rho_\Sigma$, and local score $\rho_{\widetilde{\Sigma}}$; Langevin dynamics step size $\epsilon$; {\color{red}batch size $B$}; {\color{blue}rollout length $M$}; {\color{blue}PPO clipping parameter $\epsilon_{\text{clip}}$; entropy coefficient $c_{\text{ent}}$; GAE coeff $\lambda$; discount $\gamma = e^{\beta \Delta t}$} 
    % discount factor $\gamma = e^{-\beta \Delta t}$
    \STATE Initialize neural networks:\\
    \textbf{Actor} $\Pi_{\psi_0}: \R^d \to \mathcal{P}(\R^k)$\\
    \textbf{Critic} $V_{\theta_0}: \R^d \to \R$, {\color{red} \textbf{Target} $T_{\theta_0}: \R^d \to \R$, same as Critic} \\
    \textbf{Global Score} $\Sigma_{\varphi_0}: \R^d \to \R^d$\\
    {\textbf{Local Score} $\widetilde{\Sigma}_{\xi_0}: \R^d \to \R^d$}\vspace{0.2cm}
    \STATE {\color{red}Sample $X_{t_0} = \{X_{t_0}^1, X_{t_0}^2, \cdots, X_{t_0}^B \}$, where $X_{t_0}^i \sim \xi$}\vspace{0.2cm}

    {\color{blue}\STATE Storage for Rollout of size $M$: 
    $
    \mathcal{D} = \{(X_{t_m}^{i}, A_{t_m}^{i}, r_{t_m+1}^{i}, X_{t_m+1}^{i}) \mid m=0,\ldots,M-1; i=1,\ldots,B\}
    $\vspace{0.2cm}}

   \FOR{$n=0,\dots,N-1$}%\vspace{0.2cm}
      \STATE Compute score loss for $\Sigma$:
      $\quad L_\Sigma (\varphi_n) = {\color{red} \frac{1}{B}\sum_{i=1}^B}\tr\left( \nabla_x \Sigma_{\varphi_n}(X_{t_n}^{\color{red} i}) \right) + \frac{1}{2}\norm{\Sigma_{\varphi_n}(X_{t_n}^{\color{red} i})}_2^2$%\vspace{0.2cm}
      
      \STATE Update $\Sigma$ with Adam:
      $\quad \varphi_{n+1} = \varphi_n -\rho_\Sigma \nabla_{\varphi} L_\Sigma (\varphi_n)$%\vspace{0.2cm}

      {\STATE Compute score loss for $\widetilde{\Sigma}$:
      $\quad L_{\widetilde{\Sigma}} (\xi_n) = {\color{red}\frac{1}{B} \sum_{i=1}^B}\tr\left( \nabla_x \widetilde{\Sigma}_{\xi_n}(X_{t_n}^{\color{red} i}) \right) + \frac{1}{2}\norm{\widetilde{\Sigma}_{\xi_n}(X_{t_n}^{\color{red} i})}_2^2$%\vspace{0.2cm}

      \STATE Update $\widetilde{\Sigma}$ with Adam:
      $\quad \xi_{n+1} = \xi_n -\rho_{\widetilde{\Sigma}} \nabla_{\xi} L_{\widetilde{\Sigma}} (\xi_n)$}%\vspace{0.2cm}
      
      \STATE Generate mean field samples $S_{t_n} = \left(S_{t_n}^{(1)}, \dots, S_{t_n}^{(k)}\right)$ from $\Sigma_{\varphi_{n+1}}$ and $\widetilde{S}_{t_n} = \left(\widetilde{S}_{t_n}^{(1)}, \dots, \widetilde{S}_{t_n}^{(k)}\right)$ from $\widetilde{\Sigma}_{\xi_{n+1}}$ using Langevin dynamics with step size $\epsilon$, and compute $\overline{\mu}_{S_{t_n}}$ and $\overline{\mu}_{\widetilde{S}_{t_n}}$.%\vspace{0.2cm}

      \STATE Sample action {\color{red} for i = 1, 2, ..., $B$ (vectorized)}:
      $\quad A_{t_n}^{\color{red} i} \sim \Pi_{\psi_n}(\cdot \mid X_{t_n}^{\color{red} i})$%\vspace{0.2cm}
      
      {\STATE Observe reward from environment {\color{red} for i = 1, 2, ..., $B$}:
      $\quad r_{n+1}^{\color{red} i} = -f(X_{t_n}^{\color{red} i}, \overline{\mu}_{S_{t_n}}, \overline{\mu}_{\widetilde{S}_{t_n}}, A_{t_n}^{\color{red} i}) \Delta t$}%\vspace{0.2cm}
      
      \STATE Observe next state from environment {\color{red} for i = 1, 2, ..., $B$ (vectorized)}:\\
        $\quad X_{t_{n+1}}^{\color{red} i} = b(X_{t_n}^{\color{red} i}, \overline{\mu}_{S_{t_n}}, A_{t_n}^{\color{red} i})\Delta t + \sigma(X_{t_n}^{\color{red} i}, \overline{\mu}_{S_{t_n}}, A_{t_n}^{\color{red} i}) \sqrt{\Delta t} \,z_n, \quad z_n \sim \mathcal{N}(0,1)$%\vspace{0.2cm}

      {\color{blue}\STATE Store the transition in $\mathcal{D}$:
      $\mathcal{D} \leftarrow \mathcal{D} \cup \{(X_{t_n}^{i}, A_{t_n}^{i}, r_{n+1}^{i}, X_{t_{n+1}}^{i})_{i=1}^B\}$}%\vspace{0.2cm}

      {\color{blue}\IF{$(n+1)$ \% $M = 0$} %\vspace{0.2cm}
         \STATE For $m = M-1$ down to $0$, and for each $i=1,\ldots,B$:
         \[
         \delta_m^{i} = r_{m+1}^{i} + \gamma V_{\theta_n}(X_{t_{m+1}}^{i}) - V_{\theta_n}(X_{t_m}^{i}), 
         \quad 
         \hat{A}_{t_m}^{i} = \delta_m^{i} + \gamma \lambda \hat{A}_{t_{m+1}}^{i} \quad \text{with } \hat{A}_{t_M}^{i} = 0, 
         \quad 
         \hat{R}_{t_m}^{i} = \hat{A}_{t_m}^{i} + V_{\theta_n}(X_{t_{m}}^{i})
         \]
         % \[
         % \hat{A}_{t_m}^{i} = \delta_m^{i} + \gamma \lambda \hat{A}_{t_{m+1}}^{i}, \quad \text{with } \hat{A}_{t_M}^{i} = 0
         % \]
         % \[
         % \hat{R}_{t_m}^{i} = \hat{A}_{t_m}^{i} + V_{\theta_n}(X_{t_m}^{i})
         % \]

         \STATE Compute critic loss using GAE returns:
         $
         L_V(\theta_n) = \frac{1}{B M}\sum_{m=0}^{M-1}\sum_{i=1}^B (\hat{R}_{t_m}^{i} - V_{\theta_n}(X_{t_m}^{i}))^2
         $ %\vspace{0.2cm}

         \STATE Update critic with Adam:
         $\quad \theta_{n+1} = \theta_n - \rho_V \nabla_{\theta} L_V(\theta_n)$ %\vspace{0.2cm}

         \STATE Let $\psi_{\text{old}} = \psi_n$ (old actor before update). Define the probability ratio
         $
         r_{t_m}^{i}(\psi_n) = \frac{\Pi_{\psi_n}(A_{t_m}^{i} \mid X_{t_m}^{i})}{\Pi_{\psi_{\text{old}}}(A_{t_m}^{i} \mid X_{t_m}^{i})}.
         $%\vspace{0.2cm}
         Compute actor loss using PPO with clipping and entropy regularization over the entire rollout:
         $$
         L_{\Pi}(\psi_n) = \frac{1}{B M}\sum_{m=0}^{M-1}\sum_{i=1}^B \biggl[ \min\bigl(r_{t_m}^{i}(\psi_n)\hat{A}_{t_m}^{i}, \text{clip}(r_{t_m}^{i}(\psi_n), 1-\epsilon_{\text{clip}},1+\epsilon_{\text{clip}})\hat{A}_{t_m}^{i}\bigr) + c_{\text{ent}}\mathcal{H}(\Pi_{\psi_n}(\cdot \mid X_{t_m}^{i})) \biggr]
         $$
         
         \STATE Update actor with Adam:
         $\quad \psi_{n+1} = \psi_n - \rho_{\Pi} \nabla_{\psi} L_\Pi(\psi_n)$ %\vspace{0.2cm}

         \STATE {\color{blue}\text{Clear the rollout storage: } $\mathcal{D} = \varnothing$}

      \ELSE
         \STATE {\color{blue}$\psi_{n+1} = \psi_n, \quad \theta_{n+1} = \theta_n$ (just carry forward)}
         \STATE {\color{red}}
               \vspace{-1ex}
      \ENDIF
      \STATE {\color{red}$T_{n+1} \leftarrow V_{n+1}$ if n divisible by 200, else $T_{n+1} \leftarrow T_n$}}

   \ENDFOR %\vspace{0.2cm}
   \RETURN $(\Pi_{\psi_N}, \Sigma_{\varphi_N}, \widetilde{\Sigma}_{\xi_N})$
\end{algorithmic}
\end{algorithm}

\pagebreak

\section{Experimental Results}\label{sec:experiments}

\subsection{The Linear-Quadratic Benchmark}

To compare our proposed algorithms against the baseline, we choose the linear-quadratic (LQ) control problem in the original paper  \cite{angiuli2024deepreinforcementlearninginfinite} as the evaluation benchmark. The LQ problem is formulated as the minimization of the following:
$$
\mathbb{E}\Biggl[\int_0^{\infty} e^{-\beta t}\biggl(\frac{1}{2} \alpha_t^2+c_1\left(\mathrm{X}_t^{\alpha, \mu}-c_2 m\right)^2+c_3\left(\mathrm{X}_t^{\alpha, \mu}-c_4\right)^2
        \quad {}+\tilde{c}_1\left(\mathrm{X}_t^{\alpha, \mu}-\tilde{c}_2 m^{\alpha, \mu}\right)^2+\tilde{c}_5\left(m^{\alpha, \mu}\right)^2\biggr) \mathrm{d} t\Biggr]
$$
subject to the dynamics
\begin{equation*} \label{eq: mfcg lq dynamics}
    \mathrm{d}X_t^{\alpha, \mu}=\alpha_t \,\mathrm{d} t+\sigma \, \mathrm{d}W_t, \qquad t \in [0, \infty)
\end{equation*}
where $m = \int x \, \dd \mu(x)$ and $m^{\alpha,\mu} = \int x \, \dd \mu^{\alpha, \mu}(x)$ and the fixed point condition $m=\lim _{t \rightarrow \infty} \mathbb{E}(X_t^{\hat{\alpha}, \mu})=m^{\hat{\alpha}, \mu}$ where $\hat{\alpha}$ is the optimal action. The LQ problem has an analytical solution, where we could compute the value function, optimal control, and the limiting distribution of global/local distributions in analytical form. We can therefore evaluate the success of our algorithm by comparing our learned solutions against these analytical solutions.   For brevity, we omit the analytical forms of these solutions here; they can be found on page 23 of the original paper and we kept it in appendix \ref{ana} as reference. 

\subsection{Experimental Setup}

\paragraph{LQ Benchmark} For all experiments, we construct the LQ problem using the following parameters: $c_1 = 0.5$, $c_2 = 1.5$, $c_3 = 0.5$, $c_4 = 0.25$, $\tilde{c}_1 = 0.3$, $\tilde{c}_2 = 1.25$, $\tilde{c}_5 = 0.25$, discount factor $\beta = 1$

\paragraph{Neural Network Config}
For all experiments, we used the same networks for the critic and the global/local scores: a feed-forward neural network with one hidden layer of size 128 and a Tanh activation function, The actor network is a feed-forward neural network with 1 hidden layer (with size 64 and Tanh activation function) plus two separate ``head" layers of size 64 to respectively predict the mean and standard deviation. We put a softmax layer on top of the ``head" for the standard deviation layer to ensure positive output. 

\paragraph{Hyperparameters} We used the same configuration as in the original paper \cite{angiuli2024deepreinforcementlearninginfinite}. We used $(\rho_\Pi, \rho_V, \rho_\Sigma, \rho_{\widetilde{\Sigma}}) = (5\times 10^{-6}, 10^{-5}, 10^{-6}, 5 \times 10^{-4}$). For the Langevin dynamics sampling, we use $\epsilon = 5 \times 10^{-2}$ for 200 iterations, drawing $k=1000$ samples. For all experiments, we also used learning rate scheduling as follows: For the first 10\% steps, linearly increase all learning rate 10 times; for the rest 90\% steps, linearly reduce learning rate to 25\% of original. We empirically observe that this approach seems to expedite convergence. For newly introduced hyperparameters, we use $B=8192$ for batch size. We use $M = 256, \epsilon_{\text{clip}} = 0.2, c_{\text{ent}} = 0.01, \gamma = 0.95, \beta = 1$. For the mini-batch algorithm, we used the minibatch size $b = 1024$. We train all algorithms for 200,000 steps ($N = 200,000$) except the IH-MFCG-AC-M algorithm. It is noted in the original paper that the baseline algorithm requires 2,000,000 steps to converge, which means that we expect the baseline algorithm to not converge in our experiments. For the IH-MFCG-AC-M algorithm, we empirically observed that it converges faster and we decided to train it in only 20,000 steps to show its fast convergence. 

\paragraph{Evaluation} We use the global/local score networks to draw samples of the global/local distributions and compare them against the analytical limiting distribution. We use the Critic network to compute the value function and compare it against the ground truth value function. We run the learning algorithms 5 times and compote the average along with uncertainty measures.

\subsection{Results: IH-MFCG-AC Baseline }
\begin{figure}[ht]
    \centering
    \begin{subfigure}[b]{0.6\textwidth}
        \includegraphics[width=\textwidth]{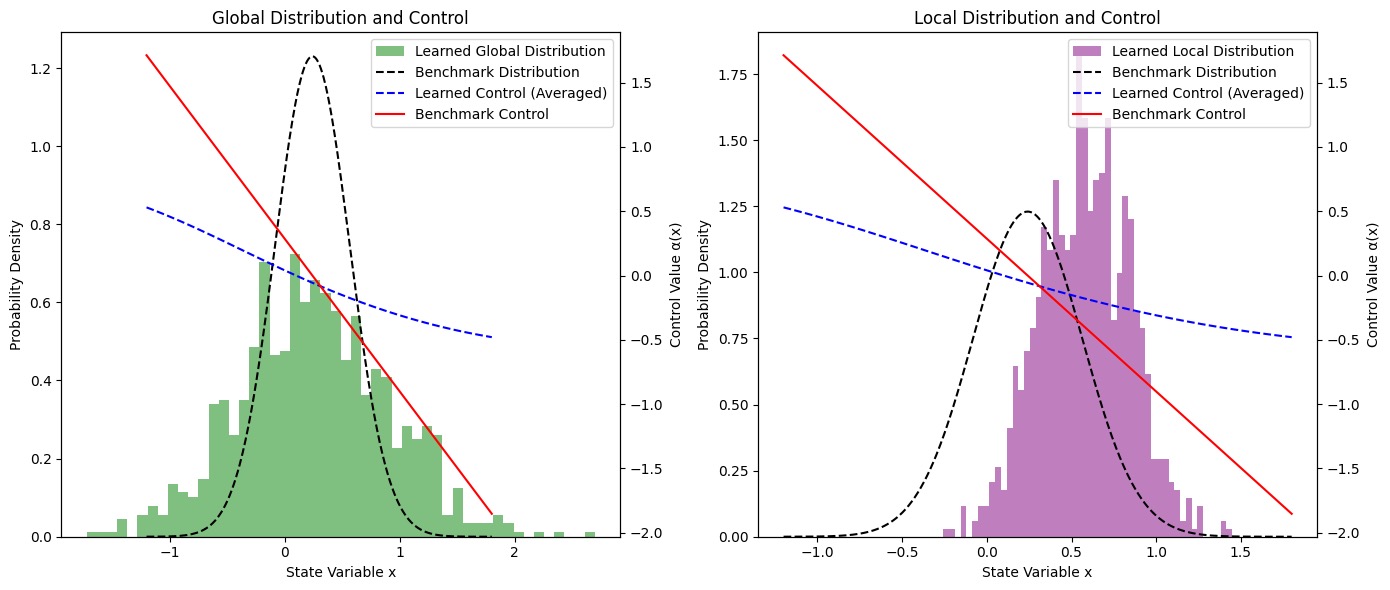}
        \caption{Learned global/local distribution vs Solution}
        \label{fig:figure1}
    \end{subfigure}
    % Second figure
    \begin{subfigure}[b]{0.35\textwidth}
        \includegraphics[width=\textwidth]{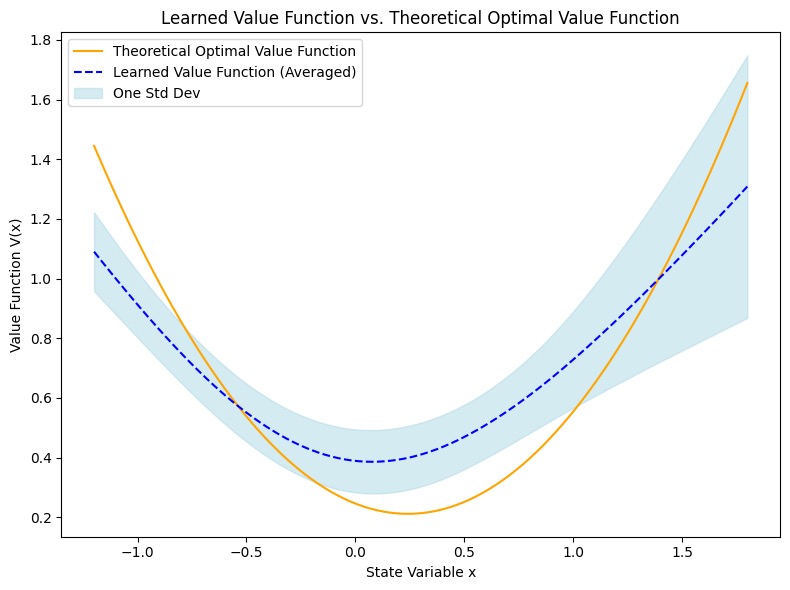}
        \caption{Learned value function vs Solution}
        \label{fig:figure2}
    \end{subfigure}
    \caption{Learned global/local distributions and value function using IH-MFCG-AC v.s. theoretical solution}
    \label{fig:baseline}
\end{figure}

Figure \ref{fig:baseline} shows the learned solutions against the analytical solutions using the baseline IH-MFCG-AC algorithm. We see that the global/local score and the value function are still pretty different from the analytical solutions. The local distribution is particularly different, shifting horizontally from the theoretical limiting distribution. This is anticipated, as the original paper reports that it takes 2,000,000 steps for the IH-MFCG-AC algorithm to converge on the LQ problem, whereas we only trained for 200,000 steps, only 10\% of the required amount. 

\subsection{Results: IH-MFCG-AC-B}
\begin{figure}[ht]
    \centering
    \begin{subfigure}[b]{0.6\textwidth}
        \includegraphics[width=\textwidth]{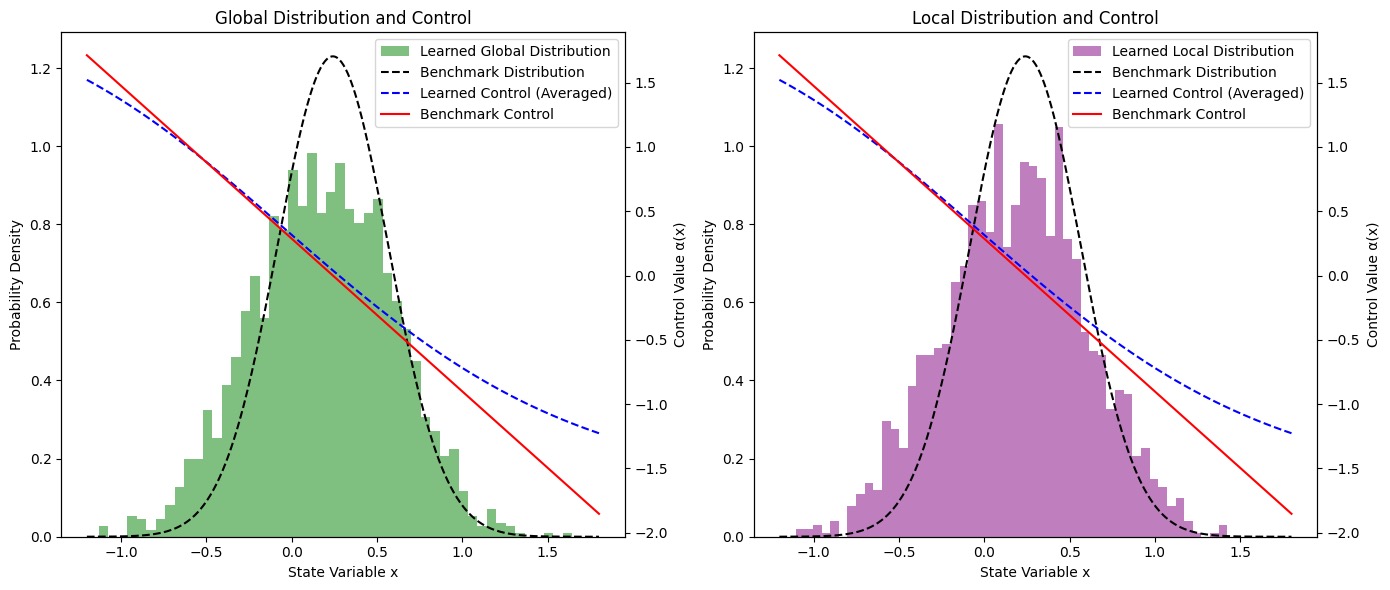}
        \caption{Learned global/local distribution vs Solution}
        % \label{fig:figure1}
    \end{subfigure}
    % Second figure
    \begin{subfigure}[b]{0.35\textwidth}
        \includegraphics[width=\textwidth]{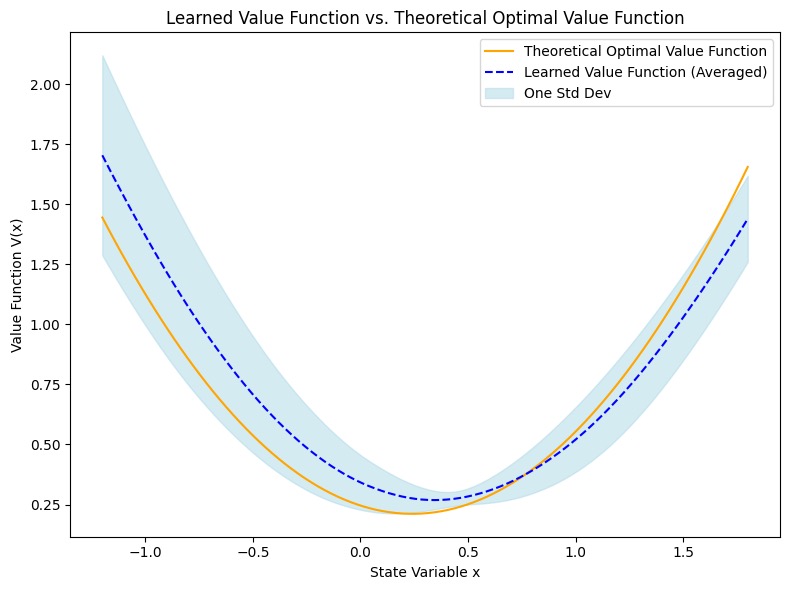}
        \caption{Learned value function vs Solution}
        % \label{fig:figure2}
    \end{subfigure}
    \caption{Learned global/local distributions and value function using IH-MFCG-AC-B v.s. theoretical solution}
    \label{fig:batch}
\end{figure}

Figure \ref{fig:batch} shows the learned solutions against the analytical solutions using the baseline IH-MFCG-AC-B algorithm. We notice that with the use of batching and the target network, The algorithm learns better and is substantially closer to the analytical solution compared to the baseline, especially on the value function and the local distribution.

\subsection{Results: IH-MFCG-AC-M}
\begin{figure}[ht]
    \centering
    \begin{subfigure}[b]{0.6\textwidth}
        \includegraphics[width=\textwidth]{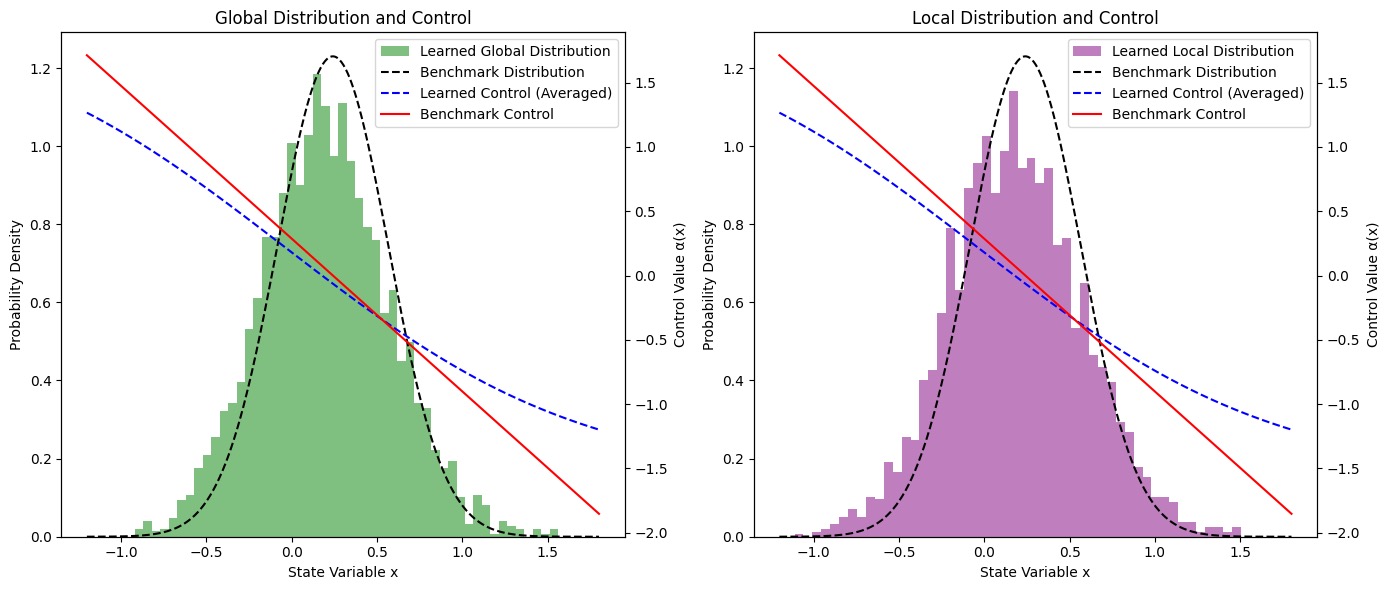}
        \caption{Learned global/local distribution vs Solution}
        % \label{fig:figure1}
    \end{subfigure}
    % Second figure
    \begin{subfigure}[b]{0.35\textwidth}
        \includegraphics[width=\textwidth]{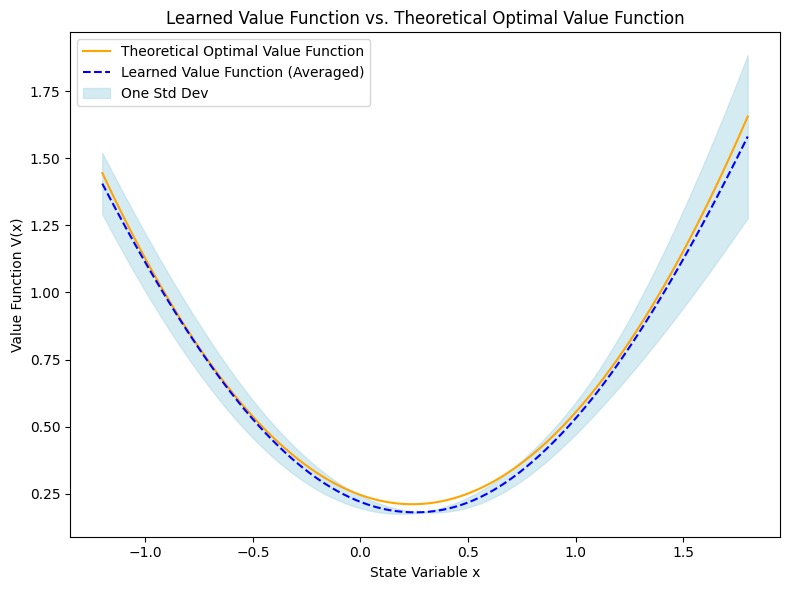}
        \caption{Learned value function vs Solution}
        % \label{fig:figure2}
    \end{subfigure}
    \caption{Learned global/local distributions and value function using IH-MFCG-AC-M v.s. theoretical solution}
    \label{fig:mini}
\end{figure}

Figure \ref{fig:mini} shows the learned solutions against the analytical solutions using the baseline IH-MFCG-AC-M algorithm. Note that we only run 20,000 steps for this algorithm, which is 10\% of the standard amount. We see that the algorithm converged very well, both in terms of local/global distribution and value function, and it achieved it with a much smaller number of steps. It is noted that the controls are slightly less well-learned compared to the IH-MFCG-AC-B experiment with 200,0000 steps, suggesting that it might benefit from slightly more training (e.g., another 10,000 steps). 

\subsection{Results: IH-MFCG-AC-DRL}
\begin{figure}[ht]
    \centering
    \begin{subfigure}[b]{0.6\textwidth}
        \includegraphics[width=\textwidth]{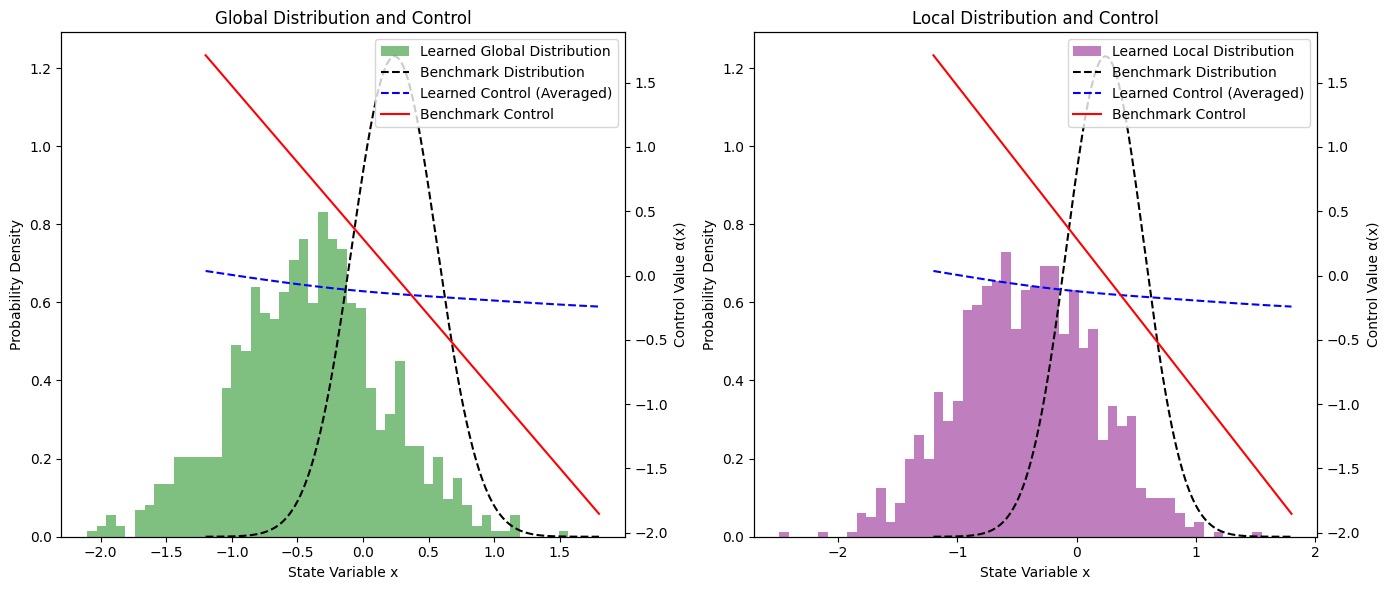}
        \caption{Learned global/local distribution vs Solution}
        % \label{fig:figure1}
    \end{subfigure}
    % Second figure
    \begin{subfigure}[b]{0.35\textwidth}
        \includegraphics[width=\textwidth]{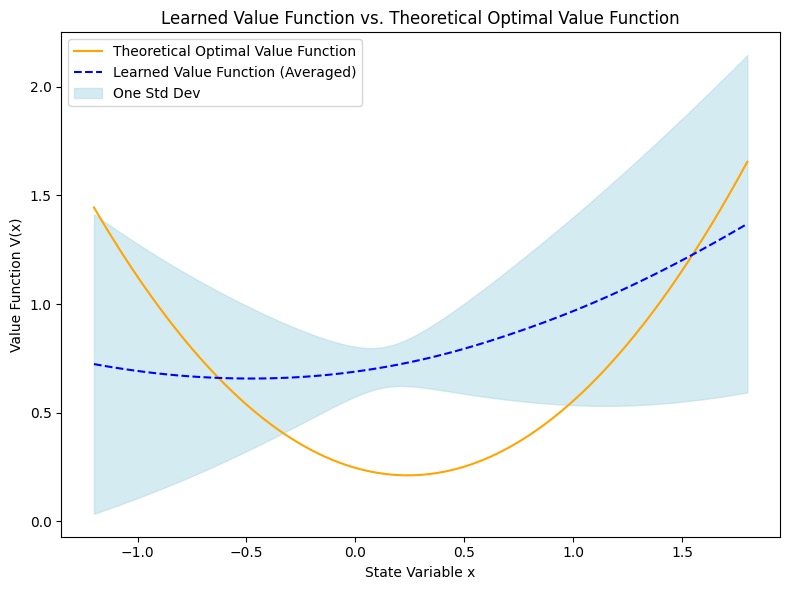}
        \caption{Learned value function vs Solution}
        % \label{fig:figure2}
    \end{subfigure}
    \caption{Learned global/local distributions and value function using IH-MFCG-AC-DRL v.s. theoretical solution}
    \label{fig:ppo}
\end{figure}

Figure \ref{fig:ppo} shows the learned solutions against the analytical solutions using the baseline IH-MFCG-AC-DRL algorithm. We see that the PPO/GAE-based algorithm did not converge at all. It is also worse than the baseline algorithm. Several discussion on the potential reason of failure is included in section \ref{expect}

\section{Discussion}\label{sec:discussion}

In this work, we have shown that deep RL methods can approximate equilibrium solutions to Mean Field Control Games (MFCGs) without directly solving coupled partial differential equations. By using standard reinforcement learning tools, we provide a scalable framework that can handle high-dimensional state spaces and complex cost structures.

\subsection{Results v.s. Expectations }

\label{expect}

In the last section, we see that the baseline IH-MFCG-AC algorithm did not converge to the theoratical solution. This is expected as we only run 10\% of the report steps required for convergence. We see that IH-MFCG-AC-B algorithm, which introduces batching, is substantially better than the baseline algorithm in terms of convergence due to better (e.g., low variance) estimation (e.g., policy gradient) at each step. In fact, we observe that the IH-MFCG-AC-B algorithm is close to convergence at 100,000 steps, at half of the standard. We record the performance of this experiment in appendix \ref{half}. IH-MFCG-AC-M converges even faster at 20,000 steps. This is surprising but not plausible, as the mini-batching procedure updates multiple times within a step, which is somewhat similar to running more total steps. 

It is surprising that IH-MFCG-AC-DRL did not perform very well. It is possible that these two algorithm are empirically proven to be successful in substantially more complicated, real-life-like tasks, and the LQ problem may be too simple for these two techniques to be successful. i.e., PPO and GAE may be a ``overkill" for the problem that we are trying to solve. Moreover, IH-MFCG-AC-DRL introduces a lot more hyperparameter (e.g., PPO clipping parameter $\epsilon_{\text{clip}}$, GAE coefficient $\lambda$, rollout length $M$, etc.), which makes hyperparameter tuning more difficult and it is likely that we have not found the proper combination. It also changes the update mechanism (i.e., we only update actor and critic once we fill the trajectory buffer, but the local/global score functions are still updated per step), which might introduce unexpected effect to the algorithm.

\subsection{Computation Runtime}

The IH-MFCG-AC-B has almost the same latency as the baseline algorithm due to vectorization and GPU acceleration, highlighting its advantage. This advantage holds for any environment that is quick to obtain responses from the current state-action pairs. In this case, the environment is vectorized, making it very fast to obtain responses for vectorized input. However, the algorithm may take more time for an environment that is not vectorized or slow to obtain responses from (e.g., complex environments that require running a costly physics engine or simulations). The IH-MFCG-AC-DRL algorithm is around twice as slow as the baseline, primarily due to the costly PPO update operation and the memory overhead introduced by the need to maintain the trajectory buffer. 

While the IH-MFCG-AC-B algorithm did not impose additional latency, the IH-MFCG-AC-B imposed latency by introducing another layer of for-loop. Suppose the cost of running batch and minibatch through the learning process is approximately the same (true for small networks with fast GPUs), then the IH-MFCG-AC-M algorithm should be $C$ times slower than the IH-MFCG-AC-B algorithm, where $C$ is the number of mini-batches. However, IH-MFCG-AC-M may be preferred over the IH-MFCG-AC-B algorithm in scenarios like (1) we can parallelize the learning on mini-batches on multiple processes/GPUs, and (2) the cost/latency of obtaining the response from the environment is high. (2) means that time is mostly spent on looping over all samples to get environmental feedback, which means the runtime will be similar between looping over all samples together (Batch) and looping over all samples piecewise (Mini-Batch).

% \subsection{Interpretation of the Results}
% Our approach leverages a representative agent who learns a policy against a mean field distribution that is itself estimated and refined during training. In principle, as the RL algorithm converges, the learned distribution and policy should approach a fixed point, reflecting an equilibrium where no agent benefits from unilaterally changing its strategy. Although our analysis has focused on simplified benchmark environments with known analytic solutions, the consistent approximation of these known equilibria suggests that the method can serve as a powerful approximation tool for more intricate settings where closed-form solutions are unavailable.

% \subsection{Connection to Theory}
% Theoretical studies of MFGs, MFCs, and MFCGs provide conditions under which unique equilibria exist. Translating these theoretical insights into practical RL algorithms remains a challenge. While our method demonstrates empirical convergence in numerical experiments, a rigorous theoretical foundation linking PDE-based equilibrium characterizations to RL-based approximations is an important area for future research. Understanding convergence rates, error bounds, and stability properties would deepen confidence in the method’s reliability and guide the selection of hyperparameters and network architectures.

\subsection{Limitations}
A number of assumptions and simplifications have facilitated our approach:
\begin{itemize}
    \item \textbf{Agent homogeneity and symmetry:} We assume a continuum of identical agents, which justifies the mean field approximation. Heterogeneous agent populations may require more sophisticated modeling.
    % \item \textbf{Markovian dynamics and full observability:} We have assumed that the state space and distribution are fully observable and Markovian. Partial observability or memory effects would complicate both the theoretical analysis and the RL implementation.
    \item \textbf{Choice of distribution approximation:} The score-matching approach provides a flexible way to represent continuous-state distributions, but it can be sensitive to hyperparameters and may require careful tuning. Alternative generative models or density estimation methods could be explored to improve robustness and scalability.
    % \item \textbf{Computational overhead:} Although batching and target networks improve stability and efficiency, the overall computational burden can still be significant, especially in high-dimensional settings or when detailed score-based distribution estimation is employed.
    \item \textbf{Theoratical Guarantee} Despiite that our algorithm outperforms baseline, it still lacks theoratical guarantee. It would be interesting to see theoretically whether our algorithm improves efficiency. 
    \item \textbf{Ablations} Our proposed algorithms contain several added component compared to the previous version. Due to the limit in compute resource and time, we did not perform an ablation study to investigate which component is most responsible for the observed performance. For example, in the IH-MFCG-AC-DRL algorithm, it is possible that GAE would improve performance, but PPO is mainly responsible for the performance downgrade. 
\end{itemize}

\subsection{Future Direction}

\paragraph{Refining Distribution Learning.} 
While we have employed a score-matching approach to represent the mean field distribution, future work might investigate other generative modeling techniques, such as normalizing flows, GANs, or VAEs, to enhance stability, reduce sensitivity to hyperparameters, and handle higher-dimensional state spaces.

\paragraph{Heterogeneous Populations and Complex Dynamics.}
Our methodology has focused on homogeneous populations and Markovian dynamics. Extending the framework to handle heterogeneous agents with differing objectives, or non-Markovian dynamics that require memory and partial observability, would broaden its applicability.

\paragraph{Integration of Model-Based and Model-Free Approaches.}
% Combining model-based techniques that leverage partial knowledge of system dynamics with model-free RL could improve convergence speed and offer theoretical guarantees. Hybrid methods might incorporate approximate PDE solutions, or analytical insights from simpler models, to guide policy updates and mitigate the risk of local minima.

We could combine model-based techniques that leverage partial knowledge of system dynamics with model-free RL. It might improve convergence speed and offer better theoretical guarantees.

\paragraph{Alternative RL Architectures and Algorithms.}
We could explore the application of other algorithms and architectures in RL on MFCGs. For example, we could try to build Q-learning variants (e.g., DQN) to solve the MFCG. For larger and more complicated tasks, we could explore the possibility of scaling up not only the sampling technique but the model itself (e.g., transformer-based networks or diffusion policies). 

% \paragraph{Real-World Applications.}
% Finally, applying this RL-based approach to complex real-world domains—such as large-scale autonomous transportation systems or multi-firm economic competition would serve as a critical test of its practical utility. 

% \subsection{Broader Implications}
% The methods developed here have the potential to impact a wide range of domains where large-scale interactions arise. In economics and finance, MFCGs model scenarios such as competition and coordination among many participants in a market. In engineering, MFCGs can inform the design of distributed control laws for large numbers of autonomous robots or vehicles. In each of these contexts, having a scalable and flexible computational tool is invaluable.

% By enabling data-driven approximations of MFCG equilibria, our approach bridges the gap between theoretical models of infinite-agent systems and their practical computational treatment. As deep RL techniques continue to evolve, incorporating model-based reasoning, multi-scale representations, or more sophisticated distribution learning methods could further enhance the accuracy and convergence properties of the proposed framework.

% In summary, our RL-based approach for MFCGs represents a significant step towards practical computations of complex mean field equilibria. While challenges remain, the work lays the foundation for more advanced techniques that blend theory and computation to tackle ever more realistic multi-agent scenarios.

\section{Conclusion and Takeaway}\label{sec:conclusion}

This paper presented a scalable, data-driven approach to solving Mean Field Control Games (MFCGs) by leveraging deep Reinforcement Learning (RL) techniques. By reformulating the MFCG problem as a Markov Decision Process and approximating both the representative agent’s policy and the population distribution, we circumvent the need to solve high-dimensional partial differential equations associated with the Hamilton-Jacobi-Bellman and Fokker-Planck systems. We adopt an algorithm from prior work and increase its efficiency and scalability by introducing batching and a target network. We name this algorithm IH-MFCG-AC-B. Building on it, we devised a new algorithm called IH-MFCG-AC-DRL, where we replaced to update the actor using PPO; used generalized advantage estimation (GAE) to estimate the advantage; and added an entropy regularizer. We also try to improve IH-MFCG-AC-B by introducing mini-batching and better trace estimation for score matching, leading to the algorithm IH-MFCG-AC-M. 

We evaluate our algorithm on a linear-quadratic benchmark where the analytical equilibrium solution is known. We observe that in resource-constrained settings (i.e., we only run 10\% steps required for the baseline algorithm to converge), the baseline algorithm did not converge. IH-MFCG-AC-B converged, and in fact it converged with even fewer steps. IH-MFCG-AC-M converges faster, requiring only 10\% of steps required for IH-MFCG-AC-B to converge (which is 1\% of steps required for IH-MFCG-AC baseline to converge). However, the IH-MFCG-AC-DRL did not converge with the given steps, possibly due to under-tuned hyperparameters and the fact that the additional techniques are usually more effective for tasks substantially more complex than the LQ benchmark. 

In summary, we successfully replicated the algorithm and results from a recent paper \cite{angiuli2024deepreinforcementlearninginfinite}, proposed several versions of the modified algorithm to improve efficiency and scalability and have seen empirical success. Our work lays the foundation for more advanced techniques to address more complicated MFCGs closely tied to real-life scenarios, such as large-scale autonomous transportation systems,  multi-firm economic competition, and inter-bank borrowing problems.

\pagebreak

% \bibliographystyle{plain}  % You can choose a different style like apalike, ieeetr, etc.
% \bibliography{references} 

\begin{thebibliography}{10}

\bibitem{achdou2012mean}
Yves Achdou, Fabio Camilli, and Italo Capuzzo-Dolcetta.
\newblock Mean field games: numerical methods for the planning problem.
\newblock {\em SIAM Journal on Control and Optimization}, 50(1):77--109, 2012.

\bibitem{achdou2020mean}
Yves Achdou and Alessio Porretta.
\newblock Mean field control problems: a probabilistic approach.
\newblock {\em Applied Mathematics \& Optimization}, 81:971--997, 2020.

\bibitem{angiuli2023reinforcementlearningalgorithmmixed}
Andrea Angiuli, Nils Detering, Jean-Pierre Fouque, Mathieu Lauriere, and Jimin Lin.
\newblock Reinforcement learning algorithm for mixed mean field control games, 2023.

\bibitem{angiuli2024deepreinforcementlearninginfinite}
Andrea Angiuli, Jean-Pierre Fouque, Ruimeng Hu, and Alan Raydan.
\newblock Deep reinforcement learning for infinite horizon mean field problems in continuous spaces, 2024.

\bibitem{busoniu2008comprehensive}
Lucian Busoniu, Robert Babuska, and Bart De~Schutter.
\newblock A comprehensive survey of multiagent reinforcement learning.
\newblock {\em IEEE Transactions on Systems, Man, and Cybernetics, Part C (Applications and Reviews)}, 38(2):156--172, 2008.

\bibitem{carmonabook}
Ren{\'e} Carmona.
\newblock {\em Lectures on BSDEs, Stochastic Control, and Stochastic Differential Games with Financial Applications}.
\newblock SIAM, 2016.

\bibitem{carmona2018probabilistic}
Ren{\'e} Carmona and Fran{\c{c}}ois Delarue.
\newblock {\em Probabilistic theory of mean field games with applications I-II}.
\newblock Springer, 2018.

\bibitem{huang2006large}
Minyi Huang, Peter~E Caines, and Roland~P Malham{\'e}.
\newblock Large-population cost-coupled lqg problems with nonuniform agents: Individual-mass behavior and decentralized $\varepsilon$-nash equilibria.
\newblock {\em IEEE Transactions on Automatic Control}, 52(9):1560--1571, 2007.

\bibitem{hyvarinen2005}
Aapo Hyv{{\"a}}rinen.
\newblock Estimation of non-normalized statistical models by score matching.
\newblock {\em Journal of Machine Learning Research}, 6(24):695--709, 2005.

\bibitem{lasry2007mean}
Jean-Michel Lasry and Pierre-Louis Lions.
\newblock Mean field games.
\newblock {\em Japanese Journal of Mathematics}, 2(1):229--260, 2007.

\bibitem{trpo}
John Schulman, Sergey Levine, Philipp Moritz, Michael~I. Jordan, and Pieter Abbeel.
\newblock Trust region policy optimization, 2017.

\bibitem{gae}
John Schulman, Philipp Moritz, Sergey Levine, Michael Jordan, and Pieter Abbeel.
\newblock High-dimensional continuous control using generalized advantage estimation, 2018.

\bibitem{ppo}
John Schulman, Filip Wolski, Prafulla Dhariwal, Alec Radford, and Oleg Klimov.
\newblock Proximal policy optimization algorithms, 2017.

\bibitem{song2019generative}
Yang Song and Stefano Ermon.
\newblock Generative modeling by estimating gradients of the data distribution.
\newblock {\em Advances in neural information processing systems}, 32, 2019.

\bibitem{reinforce}
Richard~S Sutton, David McAllester, Satinder Singh, and Yishay Mansour.
\newblock Policy gradient methods for reinforcement learning with function approximation.
\newblock In S.~Solla, T.~Leen, and K.~M\"{u}ller, editors, {\em Advances in Neural Information Processing Systems}, volume~12. MIT Press, 1999.

\bibitem{zhang2021multi}
Kaiqing Zhang, Zhuoran Yang, Tao Liu, and Tamer Başar.
\newblock Multi-agent reinforcement learning: A selective overview of theories and algorithms.
\newblock {\em Handbook of Reinforcement Learning and Control}, pages 321--384, 2021.

\end{thebibliography}

\pagebreak

\section{Appendix}

\label{appendix}

\subsection{Appendix 1: Analytical Solution of LQ problem}

\label{ana}
\underline{\textbf{Disclosure:} The following content in this subsection is adapted from the original paper \cite{angiuli2024deepreinforcementlearninginfinite}. } \vspace{2ex}

We present the analytic solution to the MFCG problem using notation consistent with the derivation in \cite{angiuli2023reinforcementlearningalgorithmmixed}. The value function is defined as
\begin{equation}
    \begin{split}
        v(x) \coloneqq \inf _{\alpha \in \mathbb{A}} \mathbb{E}\Biggl[\int_0^{\infty} e^{-\beta t}\biggl(\frac{1}{2} \alpha_t^2+c_1\left(\mathrm{X}_t^{\alpha, \mu}-c_2 m\right)^2+c_3\left(\mathrm{X}_t^{\alpha, \mu}-c_4\right)^2\\
        \quad {}+\tilde{c}_1\left(\mathrm{X}_t^{\alpha, \mu}-\tilde{c}_2 m^{\alpha, \mu}\right)^2+\tilde{c}_5\left(m^{\alpha, \mu}\right)^2\biggr) \mathrm{d} t \mid X_0 = x\Biggr].
    \end{split}
\end{equation}
The explicit formula $v(x) = \Gamma_2 x^2 + \Gamma_1 x + \Gamma_0$ can be derived as the solution to the Hamilton-Jacobi-Bellman equation where
\begin{align*}
    \Gamma_2 &=\frac{-\beta+\sqrt{\beta^2+8\left(c_1+c_3+\tilde{c}_1\right)}}{4}\\[1em]
    \Gamma_1 &= -\frac{2 \Gamma_2 c_3 c_4}{c_1\left(1-c_2\right)+\tilde{c}_1\left(1-\tilde{c}_2\right)^2+c_3+\tilde{c}_5}\\[1em]
    \Gamma_0 &= \frac{c_1 c_2^2 m^2+\left(\tilde{c}_1 \tilde{c}_2^2+\tilde{c}_5\right)\left(m^{\alpha, \mu}\right)^2+\sigma^2 \Gamma_2-\frac{1}{2} \Gamma_1^2+c_3 c_4^2}{\beta} .
\end{align*}
Then the optimal control for the MFCG is
\begin{equation}\label{eq: mfcg optimal control}
    \hat{\alpha}(x) = -(2 \Gamma_2 x + \Gamma_1).
\end{equation}
Substituting them yields the Ornstein-Uhlenbeck process
\[
    \dd X_t = -\left(2 \Gamma_2 X_t + \Gamma_1\right)\, \dd t + \sigma \, \dd W_t
\]
whose limiting distribution is
\begin{equation}
    \hat{\mu} = \mu^{\hat{\alpha}, \hat{\mu}} = \mathcal{N}\left( -\frac{\Gamma_1}{2\Gamma_2}, \frac{\sigma^2}{4 \Gamma_2} \right).
\end{equation}
We note that an equation for $\hat{m}$ and $m^{\hat{\alpha}, \hat{\mu}}$ that only depends on the running cost coefficients is
\begin{equation} \label{eq: mfcg mean}
    m \coloneqq \hat{m} = m^{\hat{\alpha}, \hat{\mu}} = \frac{c_3 c_4}{c_1\left(1-c_2\right)+\tilde{c}_1\left(1-\tilde{c}_2\right)^2+c_3+\tilde{c}_5}.
\end{equation}

\pagebreak 

\subsection{Appendix 2: IH-MFCG-AC-B on LQ benchmark with 100,000 steps}
\label{half}
The figures below shows the local/global distribution and the learned value function of the IH-MFCG-AC-B algorithm on LQ benchmark trained with 100,000 steps. This is a half of the steps compared to all other experiments, and we observe that the algorithm is close to converging into the theoratical solution. 

\begin{figure}[ht]
    \centering
    \begin{subfigure}[b]{0.6\textwidth}
        \includegraphics[width=\textwidth]{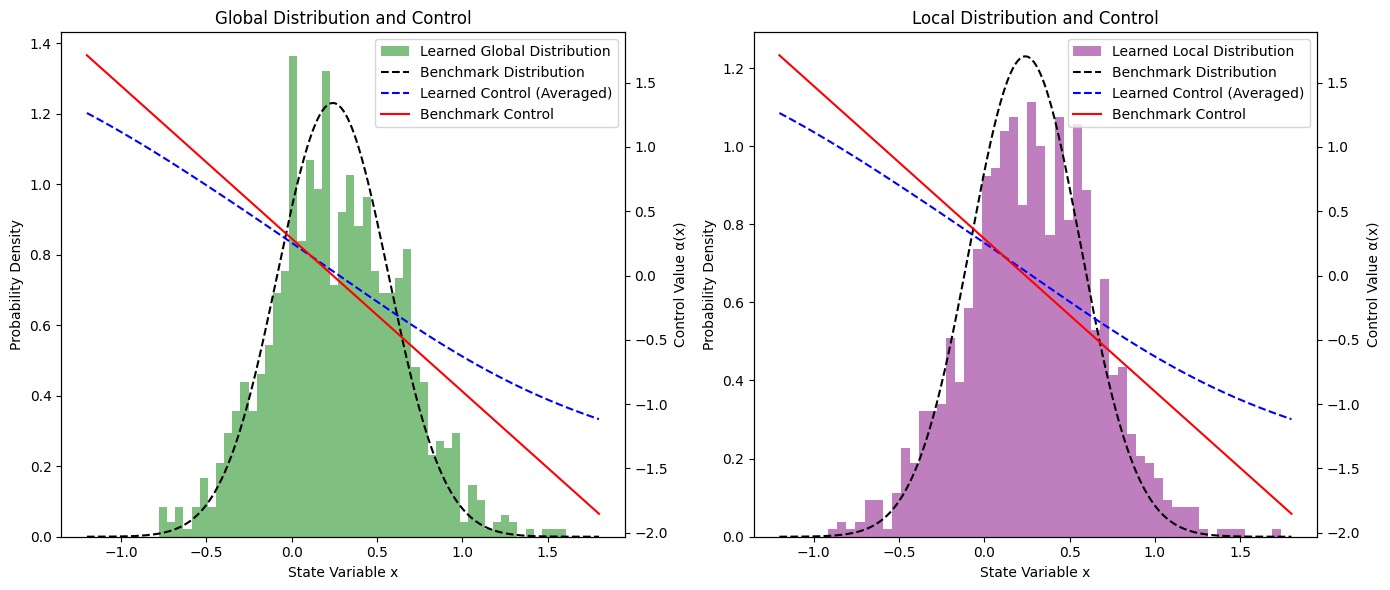}
        \caption{Learned global/local distribution vs Solution}
        % \label{fig:figure1}
    \end{subfigure}
    % Second figure
    \begin{subfigure}[b]{0.35\textwidth}
        \includegraphics[width=\textwidth]{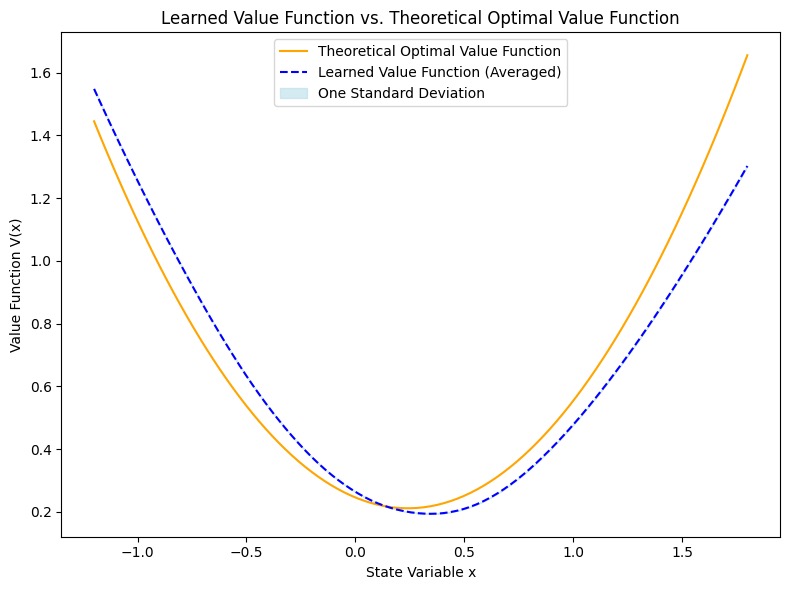}
        \caption{Learned value function vs Solution}
        % \label{fig:figure2}
    \end{subfigure}
    % \caption{Learned global/local distributions and value function using IH-MFCG-AC v.s. theoretical solution}
    % \label{fig:baseline}
\end{figure}

\pagebreak

\end{document}